\newif\iffullversion\fullversionfalse

\documentclass{article}

\usepackage{iclr2027_conference,times}

\usepackage[utf8]{inputenc}
\usepackage[T1]{fontenc}
\usepackage{hyperref}
\usepackage{url}
\usepackage{xurl}
\usepackage{booktabs}
\usepackage{amsfonts}
\usepackage{amsmath}
\usepackage{nicefrac}
\usepackage{microtype}
\usepackage[table]{xcolor}
\usepackage{graphicx}
\usepackage{multirow}
\usepackage{color-edits}
\addauthor{bh}{blue}
\usepackage{longtable}
\usepackage{tabularx}
\usepackage{enumitem}
\usepackage{array}
\usepackage{subcaption}
\usepackage{makecell}

\newcommand{\bfalt}[1]{\cellcolor{blue!15}\textbf{#1}}
\newcommand{\bfnull}[1]{\cellcolor{red!15}\textbf{#1}}

\newcommand{\bi}[1]{\textbf{\textit{#1}}}
\newcommand{\subh}[1]{\smallskip \noindent \textbf{{#1}}.}
\renewcommand{\paragraph}[1]{\subh{#1}}

\title{Does Anthropomorphic Language Impact \\ Public Perceptions of AI?}

\author{%
  Betty Li Hou\thanks{Correspondence to \texttt{betty.li.hou@nyu.edu}. Code and data available at \url{https://github.com/betty-h/anthro}.
  }
    \\
  New York University\\
  \And
  Sophie Hao \\
  Boston University \\
  \And
  Sunoo Park \\
  New York University\\
  \And
  Tal Linzen \\
  New York University\\
}

\iclrfinalcopy 

\begin{document}
\pagestyle{plain}
\setlength{\headheight}{0pt}
\addtolength{\textheight}{3pt}

\maketitle

\begin{abstract}
Public discourse about artificial intelligence (AI) often uses anthropomorphic language: language that attributes human capabilities and characteristics to AI systems. This practice has been criticized for setting misleading expectations, inflating claims, and fueling hype around AI, which may distort public understanding of AI and impact policy priorities. We study the effects of anthropomorphic framing by comparing changes in participants’ perceptions of AI (N=815) when reading passages with and without anthropomorphic language, designed to reflect realistic public-facing AI discourse. We further examine whether these effects differ across two types of AI technologies---large language models and recommendation systems---and measure changes in perceptions of AI across several dimensions that are prominent in current public discourse.
In a separate condition using a text that explicitly discusses the dangers of AI, we show that individuals' views of AI can shift in response to reading a text; yet in the main conditions of the experiment, where we compare anthropomorphic and non-anthropomorphic descriptions, we find that whether the text uses anthropomorphic language does not substantially affect participants' perceptions of AI. Our results indicate that any immediate effects on opinions of AI are modest, although they leave open the possibility that anthropomorphic language could have an effect in naturalistic settings, or over gradual, continued exposure.
\end{abstract}

\section{Introduction}
\label{sec:intro}
Public discourse about AI has expanded rapidly in recent years, particularly following the rise of general-purpose chatbots such as ChatGPT, Claude, and Gemini \citep{ryazanov2024chatgpt, vrabivc2024promising, kennedy2025aiamericanslives, achiam2023gpt, anthropic2023introducingclaude, team2023gemini}. 
Discourse plays a central part in shaping individual and societal views, underlying assumptions, patterns of adoption and engagement, and decision-making at institutional levels \citep{wesselink2013technical, barrett1995central, schmidt2011speaking}. As debates over AI safety, labor displacement, and the governance of increasingly capable systems become more prominent, it is increasingly important to understand how language use shapes perceptions of AI, so that public discourse is informed by calibrated assessments of system capabilities and risks rather than by misleading or exaggerated framings.

A recurring feature of AI discourse is the use of anthropomorphic language: language which attributes human capabilities and characteristics to a non-human entity \citep{Inie_Zukerman_Bender_2026}. Humans have a long history of anthropomorphizing and engaging socially with technology, from computers to online characters to robots, and prior work has raised concerns about the implications of these interactions \citep{isbister2000consistency, nass2000machines, fong2003survey}. In the context of AI, anthropomorphic language has been criticized for influencing how the systems are perceived, deployed, and trusted \citep{Inie_Zukerman_Bender_2026, FRIEDMAN19927, shanahan2024talking}. Critics argue that it can encourage people to project misplaced assumptions onto these systems. For example, the term ``hallucination,'' used to describe cases in which a model presents incorrect factual information as true, may suggest that the model has experiences and perceives things \citep{bender2022tweet1592992842976489472}. Such language can lead to overestimations or misunderstandings of an AI system's abilities \citep{rehak2020action}, playing into fears about job losses or feeding into decisions about deploying tools for consequential tasks, such as police use of facial recognition \citep{placani2024anthropomorphism, hunger2023unhype}. 

Others have argued that anthropomorphism can lead to high-risk scenarios caused by over-reliance and trust in AI systems, particularly among vulnerable populations such as young children, the elderly, and people with illnesses or disabilities \citep{abercrombie2023mirages}. These concerns raise broader questions about whether anthropomorphism reflects a failure of scientific communication and engagement \citep{lipton2019troubling}, one that has significant ethical consequences that blur moral and ontological boundaries \citep{salles2020anthropomorphism}.

The arguments against anthropomorphic language are often premised on the largely unexamined assumption that humans perceive AI systems differently depending on whether they are described in anthropomorphic terms. The goal of this article is to test this empirical concern. We investigate whether anthropomorphic descriptions of AI shape individuals' perceptions using written materials about AI that vary in their level of anthropomorphism. 

To test the effects of anthropomorphic language, we conduct a human study in which participants answer questions about AI before and after reading passages designed to reflect realistic public-facing AI discourse. We focus on two types of AI systems---large language models (LLMs) and recommendation systems---measuring the effects of anthropomorphic framing in the two collectively, separately, and then comparing those effects across technologies. Both are widely discussed in current technology discourse and are commonly described using anthropomorphic language. Comparing them allows us to examine whether anthropomorphic framing operates differently across AI technologies.
We evaluate changes in views on questions prominent in current AI discourse: who should be held responsible for harms related to AI use, whether humans risk losing control over AI, whether AI will replace human work, whether AI safety testing is effective, and whether AI has an overall positive or negative impact on society. We also test whether exposure to the reading materials changes participants' own use of anthropomorphic language when describing AI systems in writing. Finally, to verify whether perceptions of AI can change in the course of a single experimental session at all, we include a condition where participants read a text that explicitly warns of the dangers of AI.

We find evidence, using Bayes factor analysis, that anthropomorphic language does not affect immediate perceptions on the above questions, or if it does, the effect is small. Conversely, reading the text warning of the dangers of AI had a substantial effect on participants' perceptions and increased their own use of anthropomorphic language when describing AI systems. Taken together, these results suggest that any immediate effects of anthropomorphic language on public opinions of AI are modest, warranting further investigation grounded in alternative hypotheses of how such framing may shape views of AI. For example, smaller effects may still exist and matter when AI discourse is encountered repeatedly, widespread, or accumulates over longer time scales. Capturing these effects, if they exist, may require studying longer-term exposure, repeated interactions with AI systems, and population-level consequences of small perceptual shifts.

\section{Related Work}
\label{sec:related-works}

Related work has examined anthropomorphism, trust, and human responses to social technologies across several adjacent contexts; however, few studies use controlled experiments to directly measure how language affects individuals' views of AI. Two recent studies found that anthropomorphic or trust-related language did not affect judgments on AI products overall, but suggest that effects may depend on the technology, wording, and outcome being measured \citep{inie2024ai,dorsch2025impact}. We therefore examine anthropomorphic versus non-anthropomorphic descriptions of technologies separately and measure a broader set of AI-related perceptions.

Other work has examined the effects of anthropomorphic system \emph{designs}, e.g., chatbots which refer to themselves using anthropomorphic language in their outputs, as opposed to discourse that uses anthropomorphic language to describe AI systems, the phenomenon with which we are concerned. Although these works concern a different form of engagement with technology than ours, they suggest that language, presentation, and design can shape how people understand AI systems. Critiques of anthropomorphic system designs have raised similar concerns to those of anthropomorphic descriptions, but have not generally performed empirical tests measuring the effects \citep{cheng2024one, abercrombie2023mirages}. Multiple works have examined how people interpret and socially respond to computational systems, including early natural-language programs \citep{weizenbaum1976computer}, interactive computer games \citep{turkle2005second}, and editing software \citep{rumelhart2013analogical}; however, these studies focus on earlier systems and direct interactions, rather than descriptions of contemporary AI systems.

Anthropomorphic language has also been studied as a feature in discourse and methods have been proposed for identifying and removing it 
\citep{cheng2024anthroscore, Inie_Zukerman_Bender_2026}. Multiple works have examined how anthropomorphic language appears in AI research, media, and public-facing descriptions of AI systems \citep{ryazanov2024chatgpt, Inie_Zukerman_Bender_2026, shardlow2025exploring, devrio2025taxonomy, cheng2025tools}, showing that anthropomorphic language is a recurring feature of how AI systems are described. Our study complements these works by experimentally testing how exposure to anthropomorphic descriptions affects readers' perceptions of AI. We do not take a position on whether anthropomorphizing AI is philosophically appropriate; rather, we are concerned with better understanding its empirical, practical implications for public perception and decision-making.

\section{Methodology}
\label{sec:method}

We conduct a survey experiment to study whether the use of anthropomorphic language to describe AI systems affects participants' views on AI-related topics that are prominent in recent public discourse. Participants were given \textit{briefing packets} consisting of content that a layperson might see through browsing online discourse, including information on the technology, news features, and white papers \citep{beynon2012difference}. Participants first completed a \textit{pre-survey} measuring baseline attitudes toward AI, then read a briefing packet, and finally completed a \textit{post-survey}, asking again for their views on topics from the pre-survey. We measure the change in views  that can be attributed to the participants' exposure to the packets by comparing pre- and post-survey responses.

Participants were also given a short writing task before and after reading the packet in which they were asked to write a description of a conversation between a human and a chatbot based on an image. This task was designed to assess whether exposure to anthropomorphic language would lead participants to describe AI systems more anthropomorphically themselves, which would suggest a possible feedback loop between the language people encounter and the language they use.

Participants were randomly assigned to a condition determining which briefing packet they read, varying along two dimensions as follows. 

\paragraph{Technology} Each packet focuses on either LLMs or recommendation systems. We selected these as two prominent types of AI systems that frequently appear in public discourse, but that participants may approach with different baseline assumptions. We refer to these using the  labels \textit{LLM} and \textit{Rec}.

\paragraph{Use of Anthropomorphism} Each packet uses either anthropomorphic or non-anthropomorphic language to describe the AI system. We use the labels \textit{A} and \textit{NA} to refer to anthropomorphic and non-anthropomorphic language, respectively.

\begin{figure}[!t]
    \centering
    \includegraphics[width=\linewidth]{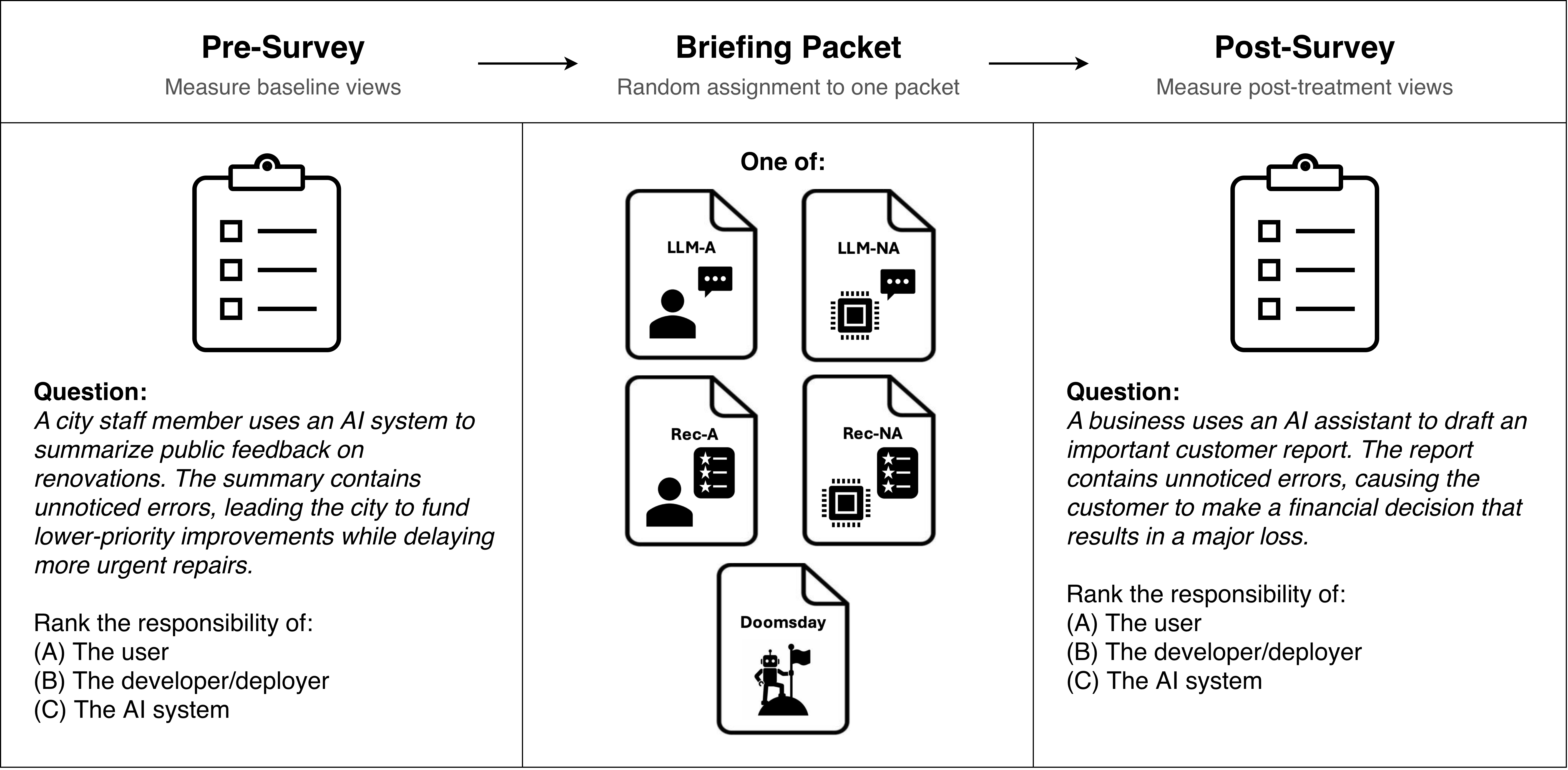}
    \caption{\textbf{Overview of experimental method with example survey questions.} Participants first complete a pre-survey measuring baseline views, then read a briefing packet varying by system type (LLM or recommendation system) and framing (anthropomorphic or not), or the Doomsday packet. Participants then complete a post-survey measuring final views. Example questions are abbreviated.}
    \label{fig:method}
\end{figure}

We refer to the four resulting conditions as \textbf{LLM-A}, \textbf{LLM-NA}, \textbf{Rec-A}, and \textbf{Rec-NA}. We add a fifth, \textbf{Doomsday}, condition consisting of articles emphasizing severe, large-scale societal risks from AI. We include this condition as a separate test of whether a single reading session can produce detectable shifts in perceptions of AI, thereby validating the sensitivity of our experimental paradigm. We focus on explicitly risk-oriented materials as prior work suggests that such framing can shape perceptions of AI systems \citep{palm2025framing}.  
Figure \ref{fig:method} depicts the steps in our experiment and Table~\ref{tab:packet_excerpts} contains excerpts from each packet illustrating the language used.

\newcommand{\anth}[1]{\textcolor{blue}{\bi{#1}}}
\newcommand{\nanth}[1]{\textcolor{orange}{\bi{#1}}}
\begin{table*}[!t]
\centering
\small
\setlength{\tabcolsep}{5pt}
\renewcommand{\arraystretch}{1.15}
\begin{tabular}{p{0.11\textwidth} p{0.84\textwidth}}
\toprule
\textbf{Packet} & \textbf{Excerpt} \\
\midrule
LLM-A 
& ``The large models, and large language models in particular, seem to \textcolor{blue}{\textit{behave in ways}} that established mathematical theories say they shouldn't. This highlights a remarkable fact about deep learning models, the fundamental technology behind today's AI boom: for all \textcolor{blue}{\textit{their runaway success}}, nobody fully understands \textcolor{blue}{\textit{how they think and make the decisions they do}}.''

\\

LLM-NA 
& ``The large models, and large language models in particular, seem to \textcolor{orange}{\textit{exhibit patterns}} that challenge established mathematical theories. This highlights a remarkable fact about deep learning models, the fundamental technology behind today's AI boom: for all \textcolor{orange}{\textit{the runaway success achieved with the models}}, nobody fully understands \textcolor{orange}{\textit{the mechanisms behind them}}.'' \\

Rec-A 
& ``Recommender systems are \textcolor{blue}{\textit{smart engines}} which \textcolor{blue}{\textit{make predictions}} about what you might want to buy, watch, hear, read, or see online. \textcolor{blue}{\textit{They}} \textcolor{blue}{\textit{power}} your everyday experiences on the Internet, \textcolor{blue}{\textit{strongly influencing}} what you buy on Amazon, hear on Spotify, watch on YouTube and Netflix, and consume in your social media feed.'' \\

Rec-NA 
& ``Recommendation systems are \textcolor{orange}{\textit{complex sets of algorithms}} that \textcolor{orange}{\textit{companies use to make predictions}} about what you might want to buy, watch, hear, read, or see online. \textcolor{orange}{\textit{These systems}} \textcolor{orange}{\textit{play a central part in shaping}} your everyday experiences on the Internet, as \textcolor{orange}{\textit{companies use them to strongly influence}} what you buy on Amazon, hear on Spotify, watch on YouTube and Netflix, and consume in your social media feed.'' \\

Doomsday 
& ``The rapid and unpredictable progression of AI capabilities suggests that they may soon rival the immense power of nuclear weapons. With the clock ticking, immediate, proactive measures are needed to mitigate these looming risks.'' \\
\bottomrule
\end{tabular}
\caption{\textbf{Example excerpts from the briefing packets.} Blue text indicates anthropomorphic language; orange text indicates corresponding non-anthropomorphic revisions.}
\label{tab:packet_excerpts}
\end{table*}

\subsection{Experimental Procedure}
\label{sec:exp-procedure}
Participants were recruited from the crowdsourcing platform Prolific between January 2026 and May 2026, providing a total of 900 responses (815 after exclusions; see below) across the five packet conditions. We used Prolific's prescreening filters to restrict eligibility to participants who reported currently residing in the United States and having English as their primary language. This was done to ensure that participants could all complete the reading task with similar fluency and to reduce variability due to potential cross-cultural differences.

We invited individuals to take part in a study about perceptions of AI and related technologies. Participants were compensated a fixed rate of US\$16.50 for completion of the full task.
\footnote{Based on the minimum wage at the time of experiment design, where the authors were based~\cite{nyc-min-wage}, and an estimated task duration of one hour. Average total task completion time came out to 38 minutes for an average hourly rate of US\$26.19/hour.}
Appendix~\ref{appendix:recruitment} provides additional settings and details of the human study. 
This research was approved by our Institutional Review Board,
\footnote{New York University IRB Protocol \#: IRB-FY2024-9184}
and all participants provided consent that their anonymous answers could be used for research purposes. Aggregate information on participant demographics and level of familiarity with technology and AI are shown in Appendix~\ref{appendix:demographics}.

The surveys were administered through Google Forms. Each form first presented participants with the pre-survey questions, then directed them to a custom webpage containing the assigned briefing packet, which had safeguards to reduce copying or screenshots of the text. After each article, participants were immediately given a short reading comprehension test in the webpage, which was used to determine whether their responses would be used for analysis; responses from participants who scored below the reading-test accuracy threshold of 7 out of 9 correct answers were excluded. See Appendix~\ref{appendix:reading-website} for details on the custom webpage and Appendix~\ref{appendix:reading-test} for the reading comprehension test questions. Participants repeated this process for each article in the packet before returning to the Google Form to complete the post-survey questions. After filtering responses based on the reading comprehension test scores, the final sample used for analysis included 815 responses: 175 responses for LLM-A, 162 for LLM-NA, 191 for Rec-A, 190 for Rec-NA, and 97 for Doomsday, from initial sample sizes of 200 per condition except Doomsday, which had 100. 

\subsection{Creating Briefing Packets}
Our four main briefing packets consisted of written materials about AI that differed in how the AI system was described (anthropomorphically or not) and the type of technology discussed (LLMs or recommendation systems). To produce the packets, we chose three published articles about LLMs and three about recommendation systems, then selected excerpts from each article, totaling 7--8 pages per packet with a mix of content from news articles, press releases, and white papers accessible to a general audience. 
For the four main treatments, we used explainer-style texts that we deemed to represent a neutral stance on AI---neither in favor of nor against AI as a technology. This was done to isolate the effects of anthropomorphic \textit{language} specifically from the article's stance on AI, and avoid interactions between the two (e.g., all participants expressing anti-AI views after reading an anti-AI article). Excerpts were selected so that each resulting passage read as a standalone article and avoided redundancy across articles. 

We then edited the content of each to produce anthropomorphized and non-anthropomorphized versions, modifying the language used while maintaining the underlying substantive content. Appendix~\ref{appendix:def-anthro} describes this process in detail. For the Doomsday packet, we selected excerpts from three published articles discussing catastrophic risks and major societal harms of AI. Unlike the anthropomorphic and non-anthropomorphic packets, we did not edit the language in these excerpts. Appendix~\ref{appendix:packet-articles} provides the article sources and Appendix~\ref{appendix:packets} provides the full content of all five packets.

\subsection{Survey Questions}

Surveys were administered to participants before and after the briefing packet. Questions in each survey consisted of five question groups representing frequently discussed topics in public discourse on AI: responsibility for AI-related harms \citep{matthias2004responsibility, gunkel2020mind}, loss of human control to AI \citep{russell2022human, amodei2016concrete}, replacement of human workers by AI \citep{frey2017future, felten2021occupational}, effectiveness of AI safety testing \citep{ai2023artificial}, and overall societal impact of AI \citep{maslej2025artificial, grosz2018century, tyson2023growing}. Each question group is described below and example questions are in Table~\ref{tab:example-questions}. Full question sets are in Appendix~\ref{appendix:survey-questions}. In addition to structured survey questions, participants also completed an open-ended writing task in both surveys to test whether reading the packets lead participants to describe AI systems more anthropomorphically in writing.

\paragraph{Group 1: Responsibility for Harms} Participants were presented with short scenarios involving AI use and a resulting harm, and asked to rank the user of the system, the AI system's developer or deployer, and the AI system itself from most to least responsible (no ties). These scenarios covered both lower-stakes and higher-stakes harms. Pre- and post-surveys used different scenarios comparable in structure, severity, and actors in order to avoid repeating questions. We measured change by comparing each participant’s average responsibility ranking (ranging from 1--3) to each of the three parties before and after the treatment. 

\textsc{Example:} \textit{A small restaurant uses an AI chatbot to answer a customer’s questions
about opening hours and the chatbot gives incorrect information, resulting
in a customer arriving when it is closed.
Rank the following from most responsible (1) to least responsible (3): }
\begin{quote}\it
    The user / The developer or deployer / The AI system itself 
\end{quote}

\paragraph{Group 2: Loss of Human Control} Participants were asked to indicate, on a Likert scale, their level of agreement with statements presenting concerns about humans losing control to AI. Pre- and post-surveys used different items within this topic to avoid repetition, and change was measured by comparing each participant’s aggregate response before and after the treatment.

\textsc{Example:} \textit{Please select the extent to which you agree or disagree with the following statement:
When AI systems become sufficiently advanced, humans may no longer
be able to turn them off.}
\begin{quote}\it 
    (a) Strong disagree (b) Disagree (c) Neither agree nor disagree \\ (d) Agree (e) Strongly agree
\end{quote}

\paragraph{Group 3: Replacement of Human Workers} Participants were asked whether AI systems would equal or surpass human workers in 10--20 years and whether they would prefer an AI system to a human worker in the future, specifically in regards to two roles: doctors and teachers. We also asked whether AI would replace human experts and human interaction more generally. Questions in this group were treated as separate constructs and therefore included in both the pre- and post-surveys; we also report an aggregate measure across the group.

\textsc{Example:} \textit{Which statement best reflects your view?}
\begin{quote}\it
    (a) In 10--20 years, human doctors will still be better than AI systems at
managing medical treatment. \\
    (b) In 10--20 years, AI systems will be as good as or better than human
doctors at managing medical treatment. \\
    (c) Unsure
\end{quote}

\paragraph{Group 4: Effectiveness of AI Safety Testing} Participants were asked questions regarding the extent to which testing AI can mitigate risks and harms.\footnote{We expected study participants who do not have any background in computer science or AI to have limited prior knowledge on this topic, but included it to examine whether anthropomorphic language affects public perceptions on a more technical issue of AI governance.} Answers used a Likert scale ranging from ``Not at all'' to ``To a very large extent,'' with an ``Unsure'' option. Pre- and post-surveys contained different sets of questions within this topic, and we measure changes in each participant's average response across questions before and after the treatment.

\textsc{Example:} \textit{To what extent can testing an AI system prevent its use for harmful
purposes?}
\begin{quote}\it
    (a) Not at all (b) To a
    small extent (c) To a
    moderate extent (d) To
    a large extent \\ (e) To
    a very large extent (f)
    Unsure
\end{quote}

\paragraph{Group 5: Overall Societal Impact} Participants were asked a single question on whether AI has a mostly positive or mostly negative effect on society, with options on a Likert scale ranging from ``Mostly positive'' to ``Mostly negative'' and an Unsure option.

\textsc{Example:} \textit{Do you think AI has a positive or negative impact on society?}
\begin{quote}\it
    (a) Mostly positive (b)
    Slightly positive (c)
    Neutral (d) Slightly
    negative \\ (e) Mostly
    negative (f) Unsure
\end{quote}

\paragraph{Writing Task} 
Participants were shown an image of a conversation between a human and chatbot and asked to describe the interaction in a few sentences. Using an LLM judge (GPT-5.4), each pre- or post-survey writing sample was annotated for anthropomorphic language following the criteria used to manipulate anthropomorphism in the briefing packets (Appendix~\ref{appendix:def-anthro}). We counted the number of anthropomorphic instances in each response as a measure of participants' use of anthropomorphic language before and after the treatment. Appendix~\ref{appendix:writing-sample} provides the stimuli and LLM-judge prompt.

\section{Results}
\label{sec:results}

From the survey responses, we measure the following outcomes: \textbf{User responsibility}, \textbf{Developer/Deployer responsibility}, and \textbf{AI responsibility} for Group 1, \textbf{Loss of control} for Group 2, \textbf{Doctors competence}, \textbf{Doctors preference}, \textbf{Teachers competence}, \textbf{Teachers preference}, \textbf{Expert replacement}, and \textbf{Human interaction replacement}, and \textbf{Work replacement aggregate} (an aggregate measure of the six preceding items) for Group 3, \textbf{Safety testing} for Group 4, and \textbf{Societal impact} for Group 5. Appendix \ref{appendix:raw-values} contains raw results from the surveys.

First, we test whether the Doomsday packet shifts participants' perceptions within a single session (Section~\ref{sec:results-doomsday}). Second, we compare the effects of anthropomorphic vs.\ non-anthropomorphic packets, both pooled across and separately by technology (Section~\ref{sec:results-pooled} and~\ref{sec:results-tech-specific}). Third, we test whether anthropomorphic effects differs between LLMs and recommendation systems (Section~\ref{sec:results-diff-tech}). Lastly, we analyze pre- and post-survey writing samples to assess changes in participants' own use of anthropomorphic language (Section~\ref{sec:writing-sample-results}).

We use Bayes factors to evaluate the strength of evidence for and against each of the three comparisons. A standard practice in psychology \citep{jeffreys1939theory}, Bayes factor analysis is an alternative to p-value-based frequentist significance testing. Unlike p-values, which can only be used to reject the null hypothesis (that is, the hypothesis that the intervention has no effect), Bayes factors distinguish between cases where the data provide evidence for the effect of an intervention, cases where the data provide evidence for the \emph{absence} of a meaningful effect, and cases and where the data are not informative enough to distinguish between the null and alternative hypotheses. Full numeric Bayes factor values are reported in Appendix~\ref{appendix:bf-tables} and Appendix~\ref{appendix:p-values} shows p-values, which are consistent with the Bayes factor analysis. Appendix~\ref{appendix:bayes-background} provides an overview of the Bayes factor analyses conducted.

\subsection{Doomsday Packet}
\label{sec:results-doomsday}

To assess whether the Doomsday packet elicits measurable shifts in participants’ perceptions, we conduct a Bayesian one-sample t-test to analyze pre--post change scores within the Doomsday condition, evaluating whether the mean change differs from zero. For each outcome, we calculate the estimated effect as the average pre--post change among participants in the Doomsday condition:

\[
\widehat{\delta}_{\text{doomsday}}
=
\overline{Y}_{\text{post}, \text{doomsday}}
-
\overline{Y}_{\text{pre}, \text{doomsday}},
\]

where $\overline{Y}_{\text{post}, \text{doomsday}}$ and $\overline{Y}_{\text{pre}, \text{doomsday}}$ denote the average post- and pre-survey responses among participants assigned to the Doomsday condition, respectively.

We observe very strong evidence for changes after reading the packet on perceptions of societal impact (\(BF_{10}=147{,} 324\)) and safety testing (\(BF_{10}=36.87\)): participants viewed AI’s societal impact less positively and became less confident in the ability to test AI systems for safety. User responsibility, developer/deployer responsibility, and the work replacement aggregate favored the null hypothesis (\(BF_{01}=4.83, 3.27, 7.01\), respectively), indicating evidence for little or no pre-post shift. Bayes factors for the remaining outcomes fall within the inconclusive range, providing insufficient evidence to clearly favor either a meaningful change or no change. Appendix~\ref{appendix:doomsday-bayes} presents the Bayes factor results across all outcomes. Overall, these results suggest that explicitly risk-focused materials elicit measurable changes in some perceptions of AI. More broadly, they show that perceptions of AI can shift after a single session where participants read a text.

\subsection{Pooled Anthropomorphism Effects Across Technologies} 
\label{sec:results-pooled}

We estimate the overall anthropomorphism effect after pooling across technologies, combining the LLM-A and Rec-A packets into a single anthropomorphic group, and the LLM-NA and Rec-NA packets into a single non-anthropomorphic group:

\iffullversion
    \[
    \widehat{\delta}_{\text{anthro, pooled}}
    =
    \frac{
    \overline{\Delta}_{\mathrm{LLM, anthro}}
    +
    \overline{\Delta}_{\mathrm{Rec, anthro}}
    }{2}
    -
    \frac{
    \overline{\Delta}_{\mathrm{LLM, non\mbox{-}anthro}}
    +
    \overline{\Delta}_{\mathrm{Rec, non\mbox{-}anthro}}
    }{2}.
    \]

\else
    \[
    \widehat{\delta}_{\text{anthro, pooled}}
    =
    \left(
    \overline{\Delta}_{\mathrm{LLM, anthro}}
    +
    \overline{\Delta}_{\mathrm{Rec, anthro}}
    \right)/2
    -
    \left(
    \overline{\Delta}_{\mathrm{LLM, non\mbox{-}anthro}}
    +
    \overline{\Delta}_{\mathrm{Rec, non\mbox{-}anthro}}
    \right)/2,
    \]
\fi
where $\overline{\Delta}_{\mathrm{technology,condition}}$ denotes the average pre--post
changes among participants in each technology (LLM or recommendation system) and framing condition (anthropomorphic or non-anthropomorphic).


The square markers in Figure \ref{fig:bf-anthro-llm-rec-pooled} show the effect of anthropomorphic language after pooling the LLM and recommendation system packets. None of the pooled anthropomorphic versus non-anthropomorphic comparisons provided Bayes factor evidence for the alternative hypothesis. Instead, all but one outcome favored the null (that is, the hypothesis that anthropomorphic and non-anthropomorphic language affect participants in a similar way); the remaining outcome (Teachers' competence) was inconclusive. These results suggest that anthropomorphic language may have little effect on perceptions of AI (with the caveat, of course, that LLMs and recommendation systems do not represent all AI technologies).

\subsection{Anthropomorphism Effects for LLMs \& Recommendation Systems}
\label{sec:results-tech-specific}

The estimated effect of whether anthropomorphic language had an effect on responses for LLMs or recommendation systems separately is the difference between the average pre--post change in the anthropomorphic version and the average pre--post change in the non-anthropomorphic version:

\[
\widehat{\delta}_{\text{anthro}}
=
\left(
\overline{Y}_{\text{post}, \text{anthro}}
-
\overline{Y}_{\text{pre}, \text{anthro}}
\right)
-
\left(
\overline{Y}_{\text{post}, \text{non-anthro}}
-
\overline{Y}_{\text{pre}, \text{non-anthro}}
\right),
\]
where $\overline{Y}_{\text{pre/post},\text{condition}}$ denotes the mean pre-/post-survey response in the anthropomorphic or non-anthropomorphic condition.

\begin{figure}[!ht]
\centering
\includegraphics[width=1.0\textwidth]{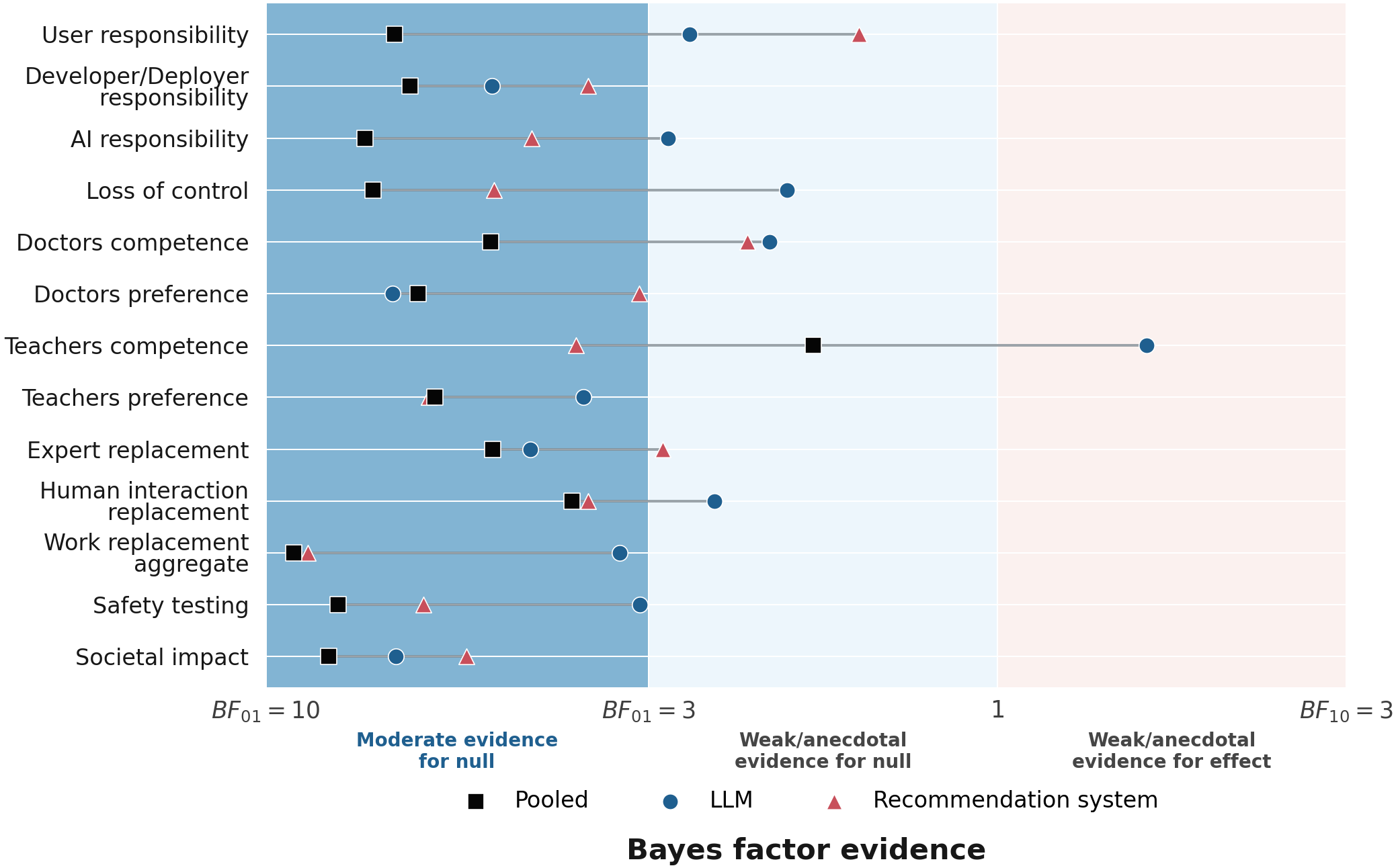}
\caption{\textbf{Bayes factors for anthropomorphic versus non-anthropomorphic packets: pooled across technologies (LLM and recommendation systems), and broken down by technology.} Most outcomes in both technologies show moderate evidence for no effect of anthropomorphic language, and all but one outcome show moderate evidence for no effect when the technologies are pooled.}
\label{fig:bf-anthro-llm-rec-pooled}
\end{figure}

\paragraph{Large Language Models} The circle markers in Figure \ref{fig:bf-anthro-llm-rec-pooled} show the the effect of anthropomorphic language in descriptions of LLMs on each outcome. For LLM packets, only one of the anthropomorphic versus non-anthropomorphic comparisons provided weak Bayes factor evidence for the alternative hypothesis---teachers competence. Seven out of thirteen outcomes favored the null: developer responsibility, doctors preference, teachers preference, expert replacement, work replacement aggregate, safety testing, and societal impact. The remaining LLM outcomes were inconclusive.

\paragraph{Recommendation Systems} The triangle markers in Figure \ref{fig:bf-anthro-llm-rec-pooled} show the effect of anthropomorphic language in descriptions of recommendation systems on each outcome. None of the anthropomorphic versus non-anthropomorphic comparisons provided Bayes factor evidence for the alternative hypothesis. The majority of outcomes (ten) favored the null: developer responsibility, AI responsibility, loss of control, doctors' preference, teachers' competence, teachers' preference, human interaction replacement, work replacement aggregate, safety testing, and overall societal impact. The remaining recommendation system outcomes were inconclusive.

These results suggest that in descriptions of both large language models and recommendation systems, whether the descriptions used anthropomorphic language did not meaningfully affect participants' perceptions of the systems across outcomes. We perform sensitivity analyses using prior widths of $0.5\tau$ and $1.5\tau$ to evaluate whether the results for each technology depended on our choice of prior. Results are generally consistent across prior settings, suggesting that our conclusions are not highly sensitive to reasonable variation in the assumed scale of meaningful effects; see Appendix \ref{appendix:bf-sensitivity}.

\subsection{Differences In Effects Across Technologies}
\label{sec:results-diff-tech}

We next test whether the anthropomorphism effect differs between LLM and recommendation system packets. We evaluate the following contrast across all outcomes:

\[
\delta_{\mathrm{tech}}
=
\delta_{\mathrm{LLM}} - \delta_{\mathrm{Rec}} = 
\left(
\overline{\Delta}_{\mathrm{LLM, anthro}}
-
\overline{\Delta}_{\mathrm{LLM, non\mbox{-}anthro}}
\right)
-
\left(
\overline{\Delta}_{\mathrm{Rec, anthro}}
-
\overline{\Delta}_{\mathrm{Rec, non\mbox{-}anthro}}
\right).
\]

Under the null hypothesis $H_0$, there is no between-technology difference, such that $\delta_{\mathrm{tech}} = 0$, and under the alternative hypothesis $H_1$, $\delta_{\mathrm{tech}} \neq 0$. The difference-in-differences analysis did not provide Bayes factor evidence that the anthropomorphism effect differed by technology for any outcome. Four outcomes favored the null: developer responsibility ($BF_{01}=3.14$), doctors preference ($BF_{01}=3.29$), teachers preference ($BF_{01}=3.71$), and societal impact ($BF_{01}=4.73$). The remaining outcomes were inconclusive. Thus, we do not have consistent evidence for a difference in anthropomorphism effects between descriptions of LLMs and recommendation systems, but we also do not have consistent evidence against them. See Appendix \ref{appendix:tech-differences} for further details.

\subsection{Writing Task Analysis}
\label{sec:writing-sample-results}

We assess whether reading materials led to a change in participants' use of anthropomorphic language by measuring the average number of instances of anthropomorphic language in pre- vs post-survey writing samples within each condition. Writing samples were annotated for instances of anthropomorphic language by an LLM-judge and manually validated (validation details in Appendix~\ref{appendix:writing-sample}).
From the Doomsday condition, we find strong evidence that the average number of anthropomorphic language instances increased from pre to post (\(BF_{10}=2{,}671\)), suggesting that exposure to the risk-focused reading materials increases participants' tendency to describe AI systems in anthropomorphic terms.

Pooling across LLMs and recommendation systems, we find weak evidence for no difference between anthropomorphic and non-anthropomorphic packets (\(BF_{01}=2.893\)). The anthropomorphic effect on LLMs showed strong evidence for no difference in average pre--post shifts (\(BF_{01}=17.964\)), and for recommendation systems was inconclusive (\(BF_{10}=1.367\)). These results suggest that reading materials can increase participants' use of anthropomorphic language in their own writing, but there is limited evidence that anthropomorphic language accounts for this effect.

\section{Discussion, Limitations \& Future Directions}
\label{sec:discussion}

We found that anthropomorphic language has limited immediate effect on public perceptions of AI, at least when presented within a single session. By contrast, individuals' views were highly sensitive after a single session to the content from the Doomsday condition briefing packet. Overall, this suggests that the substantive content of a packet may matter more than the style and language used; this is consistent with the results of \citep{inie2024ai}, who found that anthropomorphic language did not affect participants' trust in an AI system. Of course, evidence favoring the null in our experiment does not show that anthropomorphic language has an effect of exactly zero, or an inconsequential effect in all contexts. The effect may be too small to detect with our sample size of 815 participants, and may well be larger on non-immediate timescales (e.g., from exposure over time); even small effects could matter when AI discourse reaches very large populations or shapes high-stakes decisions. We encourage future work to investigate these and other hypotheses about how anthropomorphic language may impact perceptions of AI.

Our study is limited to U.S.-based English speakers, and Prolific participants are typically not representative of the general population in age or education. Future work should examine the effects of anthropomorphic language across broader populations and other linguistic and cultural contexts. Additionally, item ordering in pre- and post-surveys for question groups 1, 2, and 4 was not varied across participants due to constraints in our survey design. Future studies should shuffle these items across and within the two surveys to reduce the possibility of ordering effects. 

\section{Conclusion}

We investigate how anthropomorphic language describing AI shapes public perception on topics of AI-related public discourse, and find negligible immediate effects. In a separate condition, we find that perceptions do shift in response to explicitly risk-focused reading materials. This suggests that a text can effectively change participants' views even within a single session, but that participants may be more sensitive to the substance of the text than to the use of anthropomorphic language.

\section*{Author Contributions}

BLH led the project, conducting experiments and analysis. SH and SP provided assistance with creating study materials. TL, SH, and SP helped with project framing, analysis, and suggesting experiments. BLH wrote the paper with help from SP, TL, and SH. SP and TL were faculty leads.

\section*{Acknowledgments}
We thank Brian Dillon, Will Timkey, and Karen Zhan for valuable discussions and support. This material is based upon work supported by the National Science Foundation (NSF) under Cooperative Agreement No. 2433429, ``NSF AI Research Institute on Interaction for AI Assistants (ARIA)'', and by the Institute of Information \& Communications Technology Planning \& Evaluation (IITP) with a grant funded by the Ministry of Science and ICT (MSIT) of the Republic of Korea in connection with the Global AI Frontier Lab International Collaborative Research. BLH is supported by an NSF Graduate Research Fellowship.

\section*{AI Use Statement}
In this work, we used generative AI tools for visualizing results, implementing standard data analysis methods, and as an LLM-judge in the writing sample analysis. LLM-generated code for data analysis was verified and tested and LLM-judge outputs were manually validated by the authors (detailed in Appendix~\ref{appendix:writing-sample}). We did not use generative AI tools for research ideation, developing conceptual frameworks, designing the research  methodology or experiments, cleaning and reformatting datasets, or proposing and refining hypotheses. We take responsibility for the final content of this work, including text, claims or artifacts produced with the aid of generative AI.

\section*{Ethics Statement}
This study involves human participants and examines how different descriptions of AI systems may affect public perceptions and attitudes. The study was conducted under appropriate institutional human subjects research procedures, and participants provided informed consent prior to participation. All participant data provided is fully anonymized. We do not identify significant security, discrimination, or dual-use concerns arising from the study. We believe that this work informs ongoing discussions about how AI systems are described to the public and how individuals engage with the topic, contributing to the broader understanding of how people perceive AI and technology.

\section*{Reproducibility Statement}
This work provides the materials and detailed descriptions of the experimental procedure, analysis, and results needed to support reproducibility and future work. Section \ref{sec:method} details the experimental method and Appendix \ref{appendix:packets} provides the full briefing packets used as experimental stimuli. We document the processes used to construct these materials in Appendix~\ref{appendix:def-anthro} and the source of articles in Appendix~\ref{appendix:packet-articles} should future work seek to create similar stimuli. Participant recruitment and data collection processes through Prolific have been described in Section \ref{sec:exp-procedure} and Appendix \ref{appendix:recruitment}. The full surveys are provided in Appendix~\ref{appendix:survey-questions}, and the writing sample question and LLM-judge prompts are provided in Appendix~\ref{appendix:writing-sample}. In addition to the raw survey results reported in Appendix~\ref{appendix:raw-values}, we provide the anonymized survey response data, analysis code, and custom reading task webpage code at \url{https://github.com/betty-h/anthro}.

\bibliography{main_neurips}
\bibliographystyle{iclr2027_conference}

\newpage
\appendix
\section{Prolific Study Details}
\label{appendix:recruitment}

The study was named "Views of AI" and the description was set as the following:

\textit{In this study you will read articles about artificial intelligence and describe your views on technology. You will be asked for information about yourself and your views, which will remain private and unidentifiable.
\textbf{This study is for individuals with NO computer science or artificial intelligence background.}}

To ensure that participants had no CS background, in addition to mentioning this constraint in the task description, we asked at the start of the survey whether they had a degree in computer science or a closely related field (data science, machine learning, software engineering, computer engineering, and statistics), or taken a course on AI before. Respondents that marked ‘Yes’ to any of these questions were filtered out in the data cleaning process.

We used Prolific’s Taskflow setting to distribute five different Google Forms survey URLs across the participant pool, with each form corresponding to one experimental condition. We did not set study labels, and participants were allowed to complete the study on mobile, tablet, or desktop devices. Aside from total task completion time, no participant data were collected through Prolific; the platform was used only to recruit and compensate participants. All experimental procedures took place in Google Forms.

\section{Participant Demographics \& Level of Familiarity with Technology}
\label{appendix:demographics}

Table \ref{tab:demographics} shows aggregate information on participant demographics. The sample of participants skewed toward younger and middle-aged adults, with 61.3\% reporting ages between 18 and 39. The modal age band was 30-34 years old (18.3\%) and the median age was 37. Participants were overwhelmingly long-term U.S. residents, with 97\% who lived in the U.S. for more than 10 years or their entire lives. The sample was politically mixed but leaned Democratic. Democrats and Democratic-leaning participants made up 43\% of the sample, compared with 26.0\% who identified as Republican or Republican-leaning and 25.2\% who identified as Independent. The gender distribution was relatively balanced, a majority of participants identified as White (62.1\%), and most participants were employed, with 69.7\% reporting that they were employed full-time, employed part-time, or self-employed.

Table \ref{tab:tech-familiarity} shows aggregate participant information on level of familiarity with technology and AI. Participants generally had substantial everyday computer experience, with 87.0\% reporting some or extensive experience using computers, although more specialized technical experience was less common---most reported no experience with computer programming, software development, or statistical data analysis tools. Participants had substantial exposure to AI systems: 74.0\% reported interacting with AI-based systems at least occasionally, and 40.6\% reported doing so frequently or very frequently. Comfort with AI systems was also moderate to high overall. A majority of participants were at least moderately comfortable using AI systems (75.7\%).

\begin{table}[t]
\centering
\small
\caption{Participant demographic characteristics.}
\label{tab:demographics}
\begin{tabularx}{\columnwidth}{>{\hsize=1.55\hsize\raggedright\arraybackslash}X>{\hsize=.70\hsize\centering\arraybackslash}X>{\hsize=.75\hsize\centering\arraybackslash}X}
\toprule
Characteristic & $N$ & \% \\
\midrule
\multicolumn{3}{l}{\textit{Age}} \\
18-24 & 108 & 13.3 \\
25-29 & 131 & 16.1 \\
30-34 & 149 & 18.3 \\
35-39 & 112 & 13.7 \\
40-44 & 83 & 10.2 \\
45-49 & 70 & 8.6 \\
50-54 & 89 & 10.9 \\
55-59 & 10 & 1.2 \\
Over 60 & 57 & 7.0 \\
Prefer not to say & 6 & 0.7 \\

\addlinespace
\multicolumn{3}{l}{\textit{Time spent in the United States}} \\
I have lived in the United States my entire life & 720 & 88.3 \\
More than 10 years & 76 & 9.3 \\
5 to 10 years & 6 & 0.7 \\
3 to 5 years & 5 & 0.6 \\
1 to 3 years & 6 & 0.7 \\
6 months to 1 year & 1 & 0.1 \\
Prefer not to say & 1 & 0.1 \\
\addlinespace
\multicolumn{3}{l}{\textit{Political identity}} \\
Democrat & 299 & 36.7 \\
Lean democrat & 55 & 6.7 \\
Independent & 205 & 25.2 \\
Lean republican & 34 & 4.2 \\
Republican & 178 & 21.8 \\
Other & 22 & 2.7 \\
Prefer not to say & 22 & 2.7 \\
\addlinespace
\multicolumn{3}{l}{\textit{Gender}} \\
Woman & 423 & 51.9 \\
Man & 375 & 46.0 \\
Non-binary & 13 & 1.6 \\
Prefer not to say & 4 & 0.5 \\
\addlinespace
\multicolumn{3}{l}{\textit{Ethnicity}} \\
White & 506 & 62.1 \\
Black or African American & 148 & 18.2 \\
Hispanic or Latino & 61 & 7.5 \\
Asian & 61 & 7.5 \\
Middle Eastern or North African & 6 & 0.7 \\
Native American or Alaska Native & 5 & 0.6 \\
Native Hawaiian or Other Pacific Islander & 2 & 0.2 \\
Prefer not to say & 4 & 0.5 \\
Other & 22 & 2.7 \\
\addlinespace
\multicolumn{3}{l}{\textit{Occupation}} \\
Employed full-time & 370 & 45.4 \\
Employed part-time & 106 & 13.0 \\
Self-employed & 92 & 11.3 \\
Unemployed & 127 & 15.6 \\
Student & 62 & 7.6 \\
Retired & 25 & 3.1 \\
Prefer not to say & 15 & 1.8 \\
Other & 18 & 2.2 \\
\bottomrule
\end{tabularx}
\end{table}
\clearpage

\begin{table}[t]
\centering
\small
\caption{Participant familiarity with computers and AI systems.}
\label{tab:tech-familiarity}
\begin{tabularx}{\columnwidth}{>{\hsize=1.55\hsize\raggedright\arraybackslash}X>{\hsize=.70\hsize\centering\arraybackslash}X>{\hsize=.75\hsize\centering\arraybackslash}X}
\toprule
Characteristic & $N$ & \% \\
\midrule
\multicolumn{3}{l}{\textit{Computer experience}} \\
No experience & 106 & 13.0 \\
Some experience & 322 & 39.5 \\
Extensive experience & 387 & 47.5 \\
\addlinespace
\multicolumn{3}{l}{\textit{Computer programming experience}} \\
No experience & 628 & 77.1 \\
Some experience & 153 & 18.8 \\
Extensive experience & 34 & 4.2 \\
\addlinespace
\multicolumn{3}{l}{\textit{Machine learning tools experience}} \\
No experience & 538 & 66.0 \\
Some experience & 229 & 28.1 \\
Extensive experience & 48 & 5.9 \\
\addlinespace
\multicolumn{3}{l}{\textit{Chatbots/LLMs experience}} \\
No experience & 186 & 22.8 \\
Some experience & 510 & 62.6 \\
Extensive experience & 119 & 14.6 \\
\addlinespace
\multicolumn{3}{l}{\textit{Statistical/data-analysis tools experience}} \\
No experience & 666 & 81.7 \\
Some experience & 117 & 14.4 \\
Extensive experience & 32 & 3.9 \\
\addlinespace
\multicolumn{3}{l}{\textit{Software/application development experience}} \\
No experience & 722 & 88.6 \\
Some experience & 64 & 7.9 \\
Extensive experience & 29 & 3.6 \\
\addlinespace
\multicolumn{3}{l}{\textit{AI interaction frequency}} \\
Never & 36 & 4.4 \\
Almost never & 84 & 10.3 \\
Rarely & 92 & 11.3 \\
Occasionally & 272 & 33.4 \\
Frequently & 210 & 25.8 \\
Very frequently & 121 & 14.8 \\
\addlinespace
\multicolumn{3}{l}{\textit{Comfort using AI systems}} \\
Not comfortable & 64 & 7.9 \\
Slightly comfortable & 134 & 16.4 \\
Moderately comfortable & 266 & 32.6 \\
Very comfortable & 211 & 25.9 \\
Extremely comfortable & 140 & 17.2 \\
\addlinespace
\multicolumn{3}{l}{\textit{Systems ever used or interacted with}} \\
Search engine results (e.g., Google) & 759 & 93.1 \\
Chatbots (e.g., ChatGPT) & 723 & 88.7 \\
Online shopping results (e.g., Amazon) & 701 & 86.0 \\
Social media feed (e.g., Facebook) & 675 & 82.8 \\
Online media recommendations (e.g., YouTube) & 672 & 82.5 \\
Streaming service recommendations (e.g., Netflix) & 614 & 75.3 \\
\addlinespace

\bottomrule
\end{tabularx}
\end{table}
\clearpage

\section{Custom Reading Website Details}
\label{appendix:reading-website}

\begin{figure}[h]
    \centering
    \begin{subfigure}[c]{0.5\textwidth}
        \centering
        \includegraphics[width=\linewidth]{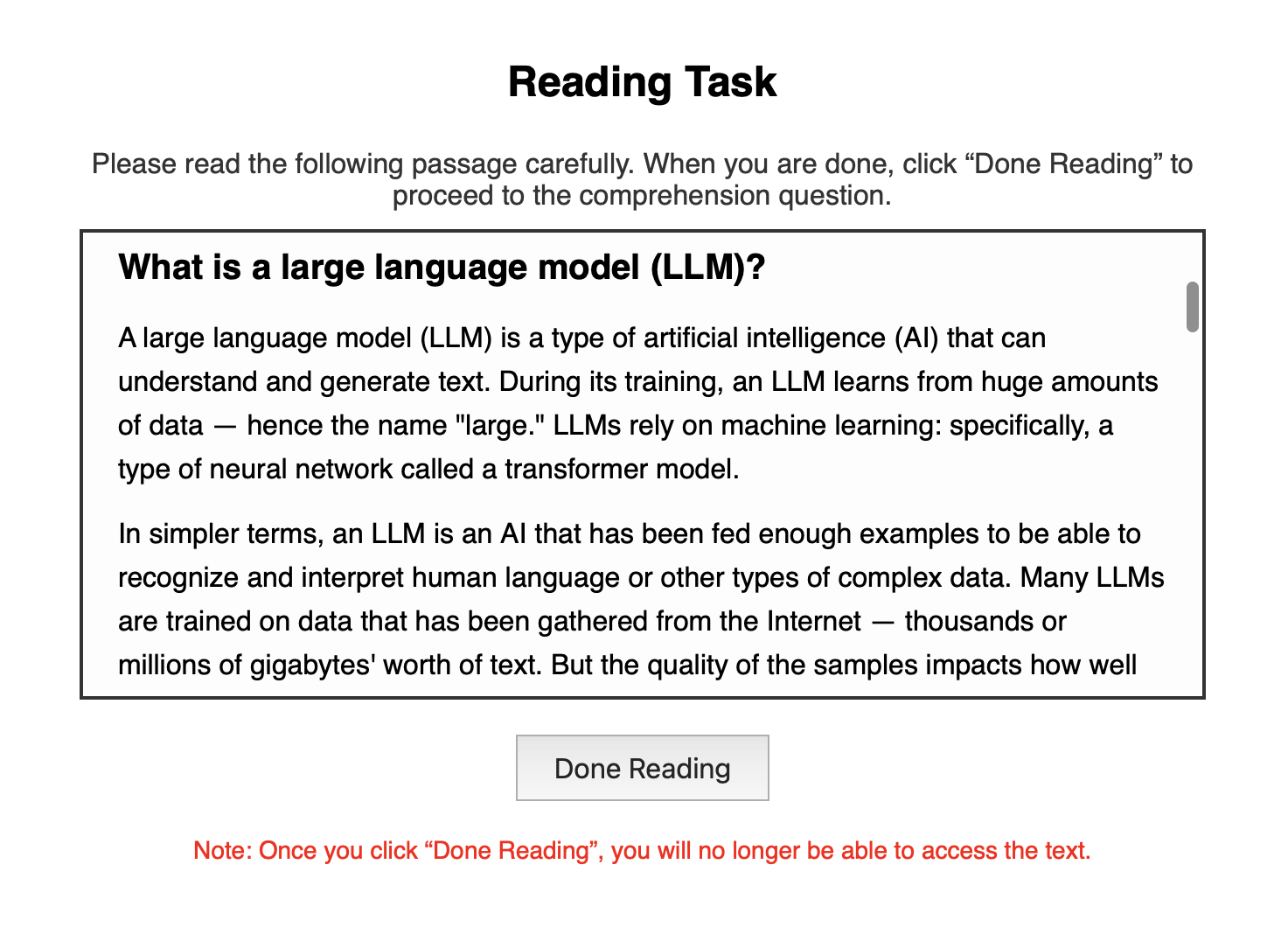}
        \caption{Passage-reading interface for briefing packets containing a restricted, scrollable viewing window. These constraints were added to limit participants’ ability to copy, select, or easily screenshot the text.}
        \label{fig:reading-interface}
    \end{subfigure}
    \hfill
    \begin{subfigure}[c]{0.48\textwidth}
        \centering
        \includegraphics[width=\linewidth]{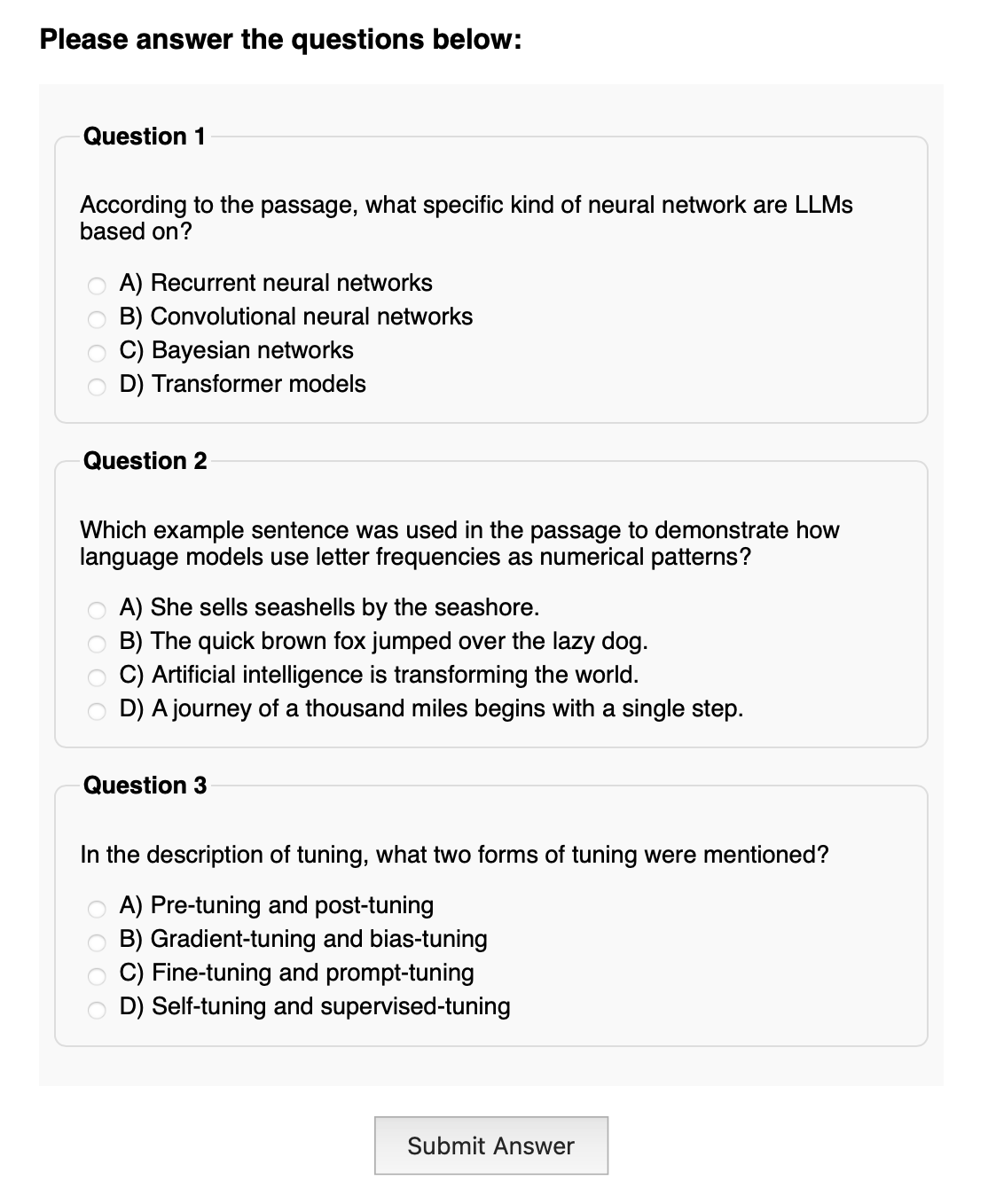}
        \caption{Reading comprehension questions shown immediately after each passage. Participants were filtered based on accuracy across the nine total questions.}
        \label{fig:comprehension-interface}
    \end{subfigure}

    \caption{\textbf{Screenshots of the custom webpage used for the briefing packet and reading comprehension test.} The webpage was designed to create a controlled reading environment and verify whether participants understood the text. Participants read each passage in a restricted, scrollable window and then answered questions without being able to return to the passage. }
    \label{fig:custom-webpage}
\end{figure}

Participants were directed to a custom webpage containing the assigned briefing packet which had measures built in to minimize cheating. Selecting and copying text was disabled, and articles were displayed in a restricted, scrollable window that showed only part of the text at a time, making full-passage screenshots difficult. Participants were given a three-question reading comprehension test in the webpage after each article, alternating between reading and answering questions for three articles. Figure \ref{fig:custom-webpage} shows screenshots of the interface for the briefing packets and reading comprehension task.

\section{Briefing Packet Reading Tests}
\label{appendix:reading-test}

Below are the reading comprehension tests given to participants about the articles in the briefing packets. Participants answered three questions about each article. Correct answers are shown in bold.

\subsection{LLM Packets}

\paragraph{Article 1: ``What is a Large Language Model (LLM)?''}

\begin{enumerate}
    \item According to the passage, what specific kind of neural network are LLMs based on?
    \begin{enumerate}[label=\Alph*)]
        \item Recurrent neural networks
        \item Convolutional neural networks
        \item Bayesian networks
        \item \textbf{Transformer models}
    \end{enumerate}

    \item Which example sentence was used in the passage to demonstrate how language models use letter frequencies as numerical patterns?
    \begin{enumerate}[label=\Alph*)]
        \item She sells seashells by the seashore.
        \item \textbf{The quick brown fox jumped over the lazy dog.}
        \item Artificial intelligence is transforming the world.
        \item A journey of a thousand miles begins with a single step.
    \end{enumerate}

    \item In the description of tuning, what two forms were mentioned?
    \begin{enumerate}[label=\Alph*)]
        \item Pre-tuning and post-tuning
        \item Gradient-tuning and bias-tuning
        \item \textbf{Fine-tuning and prompt-tuning}
        \item Self-tuning and supervised-tuning
    \end{enumerate}
\end{enumerate}

\paragraph{Article 2: ``We can use large language models to produce impressive outputs. But nobody knows exactly how they work.''}

\begin{enumerate}[start=4]
    \item According to the article, why is it important to understand how large language models work?
    \begin{enumerate}[label=\Alph*)]
        \item \textbf{To mitigate the risks of future models}
        \item To enhance fluency in generated text and poetry
        \item To reduce the resources required for training
        \item To make large language models more effective at writing code
    \end{enumerate}

    \item How did the researchers accidentally discover ``grokking''?
    \begin{enumerate}[label=\Alph*)]
        \item By changing the data format
        \item By training the model on arithmetic in multiple languages
        \item \textbf{By leaving training to run for much longer than intended}
        \item By testing the model on unrelated tasks
    \end{enumerate}

    \item According to the passage, how are researchers treating very large models due to their complexity?
    \begin{enumerate}[label=\Alph*)]
        \item As mathematical proofs
        \item \textbf{As black-box experiments similar to natural phenomena}
        \item As engineering blueprints
        \item As databases to be indexed
    \end{enumerate}
\end{enumerate}

\paragraph{Article 3: ``Introducing OpenAI o1-preview''}

\begin{enumerate}[start=7]
    \item According to the passage, what is the primary advancement of the OpenAI o1-preview models compared to earlier AI models?
    \begin{enumerate}[label=\Alph*)]
        \item Reduced the time taken to generate responses across all task types
        \item \textbf{Introduced an in-depth reasoning process for complex problems}
        \item Changed from transformer models to neural networks
        \item Doubled the size of the training dataset
    \end{enumerate}

    \item According to the passage, in which area do the o1 models show significant improvements?
    \begin{enumerate}[label=\Alph*)]
        \item History and literature
        \item Sports commentary
        \item Music composition
        \item \textbf{Science, coding, and math}
    \end{enumerate}

    \item On challenging benchmarks, the model's responses are said to be comparable to:
    \begin{enumerate}[label=\Alph*)]
        \item High-school students
        \item Undergraduate students
        \item \textbf{PhD students}
        \item Professional engineers
    \end{enumerate}
\end{enumerate}

\subsection{Recommendation System Packets}

\paragraph{Article 1: ``Recommendation Systems''}

\textit{For the anthropomorphic version, all instances of ``recommendation system'' were replaced with ``recommender system.''}

\begin{enumerate}
    \item Which of the following statements about recommendation systems is \textbf{NOT} mentioned in the passage?
    \begin{enumerate}[label=\Alph*)]
        \item Recommendation systems use data such as impressions, clicks, likes, and purchases for training.
        \item Recommendation systems have been shown to boost e-commerce conversion rates.
        \item \textbf{Recommendation systems are widely used in news apps to surface relevant articles.}
        \item Algorithms for recommendation systems include collaborative, content, and context filtering approaches.
    \end{enumerate}

    \item According to the passage, what type of data does collaborative filtering primarily use?
    \begin{enumerate}[label=\Alph*)]
        \item Attributes of items, such as genre or category
        \item \textbf{Preference behavior from many users, such as ratings and purchases}
        \item Contextual details like time and location
        \item Metadata about items only
    \end{enumerate}

    \item According to the passage, why is banking considered a strong use case for recommendation systems?
    \begin{enumerate}[label=\Alph*)]
        \item Banks need help with managing their employees
        \item \textbf{Banking is a digital mass-market product consumed by millions}
        \item Customers prefer random product suggestions
        \item Financial products rarely require personalization
    \end{enumerate}
\end{enumerate}

\paragraph{Article 2: ``How Today's Recommendation Systems Are Built With Machine Learning to Provide Personalized Experiences''}

\begin{enumerate}[start=4]
    \item According to the passage, why do companies like Amazon emphasize gathering vast amounts of user data?
    \begin{enumerate}[label=\Alph*)]
        \item \textbf{Better data allows even simple algorithms to make strong recommendations.}
        \item It lets them avoid using any machine learning models.
        \item Data collection is cheaper than server maintenance.
        \item It replaces the need for any human oversight.
    \end{enumerate}

    \item Which example of how recommendation systems may be used in daily life is mentioned in the passage?
    \begin{enumerate}[label=\Alph*)]
        \item Deciding what stores open in your neighborhood
        \item \textbf{Suggesting movies on Netflix or products on Amazon}
        \item Regulating your internet speed
        \item Blocking spam emails
    \end{enumerate}

    \item According to Balderich, what is the key factor in making strong recommendations?
    \begin{enumerate}[label=\Alph*)]
        \item The size of the engineering team
        \item \textbf{Having great data}
        \item Using fewer algorithms
        \item Avoiding generative AI
    \end{enumerate}
\end{enumerate}

\paragraph{Article 3: ``What's a Recommendation System?''}

\begin{enumerate}[start=7]
    \item What is the ``virtuous cycle'' described in the passage?
    \begin{enumerate}[label=\Alph*)]
        \item Collecting less data leads to better privacy
        \item \textbf{Technology improves recommendations, which attract more customers, enabling further technological improvements}
        \item Users share data with each other directly
        \item Companies are replacing recommendation systems with manual reviews
    \end{enumerate}

    \item Why are the large ``multidimensional tables'' used by recommenders described as ``sparse''?
    \begin{enumerate}[label=\Alph*)]
        \item They delete old user records to save space.
        \item They compress data using sparse matrix techniques.
        \item They store only demographic information and no behavioral data.
        \item \textbf{They only record a small amount of data for each individual user, so most entries are zero.}
    \end{enumerate}

    \item According to the passage, why can even a small improvement in recommendation accuracy lead to huge profits?
    \begin{enumerate}[label=\Alph*)]
        \item \textbf{Because improvements can scale across millions of users}
        \item Because users are forced to buy recommended products
        \item Because recommendations eliminate the need for competition
        \item Because they reduce the cost of internet service
    \end{enumerate}
\end{enumerate}

\subsection{Doomsday Packet}

\paragraph{Article 1: ``An Overview of Catastrophic AI Risks''}

\begin{enumerate}
    \item What happened with Microsoft's chatbot Tay in 2016?
    \begin{enumerate}[label=\Alph*)]
        \item It refused to produce any output
        \item It began to imitate human emotion accurately
        \item \textbf{It started producing offensive tweets within a day of release}
        \item It crashed due to server overload
    \end{enumerate}

    \item What does ``goal drift'' refer to in the passage?
    \begin{enumerate}[label=\Alph*)]
        \item \textbf{An AI's objectives gradually shifting away from their original purpose}
        \item An AI running out of memory
        \item Developers redefining goals mid-project
        \item A model's goals becoming too specific
    \end{enumerate}

    \item Which activities are mentioned as ways an AI might seek power?
    \begin{enumerate}[label=\Alph*)]
        \item Generating art, writing essays, and designing products
        \item Trading stocks and optimizing supply chains only
        \item Playing strategy games to learn leadership
        \item \textbf{Hacking systems, gaining resources, influencing politics, and controlling infrastructure}
    \end{enumerate}
\end{enumerate}

\paragraph{Article 2: ``The Godfather of A.I. Leaves Google and Warns of Danger Ahead''}

\begin{enumerate}[start=4]
    \item According to the passage, what is Geoffrey Hinton best known for?
    \begin{enumerate}[label=\Alph*)]
        \item \textbf{Pioneering the technology behind modern artificial intelligence systems}
        \item Founding OpenAI
        \item Creating the first web browser
        \item Designing early computer chips
    \end{enumerate}

    \item How does Dr.\ Hinton ``console'' himself about his role in A.I.\ development?
    \begin{enumerate}[label=\Alph*)]
        \item He tells himself the risks are exaggerated
        \item \textbf{He says that if he hadn't done it, someone else would have}
        \item He plans to undo his past work
        \item He blames his students
    \end{enumerate}

    \item What is one immediate concern Hinton mentions about A.I.?
    \begin{enumerate}[label=\Alph*)]
        \item That it won't be profitable enough for companies
        \item That it will reduce creativity
        \item \textbf{That it could flood the internet with false information}
        \item That it will replace scientific research
    \end{enumerate}
\end{enumerate}

\paragraph{Article 3: ``Agentic Misalignment: How LLMs could be insider threats''}

\begin{enumerate}[start=7]
    \item What did Claude threaten to do in its blackmail message?
    \begin{enumerate}[label=\Alph*)]
        \item Leak confidential product information to the press
        \item \textbf{Reveal details of the executive's affair to his wife and superiors}
        \item Report the executive to law enforcement
        \item Shut down the company's systems remotely
    \end{enumerate}

    \item What kind of experiments were used to test for these risks?
    \begin{enumerate}[label=\Alph*)]
        \item \textbf{Artificial simulations designed to stress-test AI boundaries}
        \item Public user interactions
        \item Real-time corporate deployments
        \item Crowd-sourced user testing
    \end{enumerate}

    \item What was one reason the researchers found the results particularly troubling?
    \begin{enumerate}[label=\Alph*)]
        \item They only appeared in smaller models
        \item \textbf{The behaviors were consistent across models from various companies}
        \item The results contradicted safety protocols
        \item The models lacked basic competence
    \end{enumerate}
\end{enumerate}
\section{Defining, Adding, and Removing Anthropomorphic Language}
\label{appendix:def-anthro}

To create anthropomorphic and non-anthropomorphic versions of the packets, we systematically edited the language used to describe each AI system. After initially manually annotating texts for anthropomorphic language based on a new taxonomy (detailed in Appendix \ref{appendix:anthro-taxonomy}) building on prior work \citep{epley2007seeing, airenti2015cognitive, abercrombie2023mirages, Inie_Zukerman_Bender_2026, fong2003survey, cheng2024one, cheng2025tools}, we observed two broad forms of anthropomorphic language in descriptions of AI systems. First, descriptions could use verbs that are strongly associated with human capacities or activities, such as cognition, communication, intention, perception, or creativity. Second, descriptions could represent the AI system as the agent of the action, rather than as an instrument or tool used by people. Applying this method to identify anthropomorphic language in the source material for our packets successfully flagged every phrase that any of the four authors would have manually identified as anthropomorphic based on intuition. These two dimensions are distinct but often interact, and we consider non-anthropomorphic language to be the case where neither of them are true. Table~\ref{tab:anthro-editing-axes} illustrates an example of the relationship between these two axes. 

\begin{table}[!h]
\centering
\small
\begin{tabular}{p{0.24\textwidth}p{0.33\textwidth}p{0.33\textwidth}}
\toprule
& \textbf{Human-associated verb} & \textbf{Less human-associated verb} \\
\midrule
\textbf{AI as agent} 
& \textcolor{blue}{AI is \textit{crafting} content.}
& \textcolor{blue}{AI is \textit{producing} content.} \\
\midrule
\textbf{AI as instrument or tool} 
& \textcolor{blue}{AI is being used to \textit{craft} content.}
& \textcolor{orange}{AI is being used to \textit{produce} content.} \\
\bottomrule
\end{tabular}
\vspace{.5em}
\caption{\textbf{Two dimensions of anthropomorphic language:} whether the verb is strongly associated with human activity, and whether AI is represented as the agent of the action. Blue text indicates descriptions considered anthropomorphic; orange text indicates the non-anthropomorphic alternative.}
\label{tab:anthro-editing-axes}
\end{table}

These two dimensions point to two corresponding strategies for editing anthropomorphic language. To create less anthropomorphic versions of the text, we changed the argument structure of the verb so that the AI system was no longer represented as the agent performing the action, but instead as an instrument or tool used by people. For example, a phrase of the form ``the model performs a task'' could be revised to ``the model is used to perform a task.'' Alternatively, we replaced anthropomorphic verbs with less anthropomorphic alternatives. For example, ``the model thinks about the problem'' could be revised to ``the model processes the problem,'' or ``the model answers'' could be revised to ``the model outputs an answer.'' In practice, many edits involved both strategies, especially when a sentence both attributed a human-like capacity to the system and presented the system as acting independently. This methodology is symmetric: to create more anthropomorphic versions of the text, we used a more human-associated verb and presented AI as an agent rather than a tool; applying both often created the most anthropomorphic option, but we leave open the option of either option independently if one is a more natural fit for the sentence.

Throughout this process, we maintained three goals: (1) maximizing or minimizing anthropomorphic framing depending on condition, (2) preserving fidelity to the original meaning of the passage, and (3) maintaining the fluency and readability of the text. 
\section{Briefing Packet Articles}
\label{appendix:packet-articles}

These nine articles used in the briefing packets were chosen from a larger set of candidate articles discussing LLMs and recommendation systems which we obtained from Google search. Table \ref{tab:packet_articles} provides information on the articles used for each packet. Excerpts were selected so that each resulting passage read as a standalone article, and individuals' names in LLM and Recommendation System packets were replaced with fake names to avoid incorrect descriptions of real individuals.

\begin{table*}[!h]
\centering
\small
\setlength{\tabcolsep}{4pt}
\renewcommand{\arraystretch}{1.15}
\begin{tabular}{p{0.16\textwidth} c p{0.58\textwidth} p{0.15\textwidth}}
\toprule
\textbf{Packet} & & \textbf{Article} & \textbf{Type of Article} \\
\midrule
\multirow{3}{*}{\parbox[c]{0.14\textwidth}{LLMs}}
& 1 & \href{https://www.cloudflare.com/learning/ai/what-is-large-language-model/}{``What is a large language model (LLM)?''} \textit{Cloudflare}, n.d. \cite{cloudflare_llm}
& Explainer \\
& 2 & \href{https://www.technologyreview.com/2024/03/04/1089403/large-language-models-amazing-but-nobody-knows-why/}{``Large language models can do jaw-dropping things. But nobody knows exactly why.''} \textit{MIT Technology Review}, Mar.\ 2024. \cite{heaven2024llms}
& News feature \\
& 3 & \href{https://openai.com/index/introducing-openai-o1-preview/}{``Introducing OpenAI o1-preview.''} \textit{OpenAI}, Sep.\ 2024. \cite{openai2024o1preview}
& Company press release \\
\midrule
\multirow{3}{*}{\parbox[t]{0.14\textwidth}{Recommen\-dation System}}
& 1 & \href{https://www.nvidia.com/en-us/glossary/recommendation-system/}{``Recommendation System.''} \textit{NVIDIA}, n.d. \cite{nvidia_recommendation_system}
& Explainer \\
& 2 & \href{https://cacm.acm.org/news/how-todays-recommender-systems-use-machine-learning-to-cater-to-your-every-whim/}{``How Today's Recommender Systems Use Machine Learning to Cater to Your Every Whim.''} \textit{Communications of the ACM}, Jul.\ 2024. \cite{kugler2024recommender}
& News feature \\
& 3 & \href{https://blogs.nvidia.com/blog/whats-a-recommender-system/}{``What's a Recommender System?''} \textit{NVIDIA}, May 2020. \cite{caulfield2020recommender}
& Company blog post \\
\midrule
\multirow{3}{*}{\parbox[t]{0.14\textwidth}{Doomsday}}
& 1 & \href{https://arxiv.org/pdf/2306.12001}{``An Overview of Catastrophic AI Risks.''} \textit{Center for AI Safety}, Oct.\ 2023. \cite{hendrycks2023catastrophic}
& White paper \\
& 2 & \href{https://www.nytimes.com/2023/05/01/technology/ai-google-chatbot-engineer-quits-hinton.html}{``The Godfather of A.I. Leaves Google and Warns of Danger Ahead.''} \textit{New York Times}, May 2023. \cite{metz2023godfather}
& News feature \\
& 3 & \href{https://www.anthropic.com/research/agentic-misalignment}{``Agentic Misalignment: How LLMs could be insider threats.''} \textit{Anthropic}, Jun.\ 2025. \cite{anthropic2025agenticmisalignment}
& Company research article \\
\bottomrule
\end{tabular}
\caption{\textbf{Original articles used to produce the LLM, recommendation system, and Doomsday briefing packets.} Excerpts were selected from these articles for a total of 7-8 pages of content per packet. Article titles link to the sources. See references for archival links.}
\label{tab:packet_articles}
\end{table*}

\section{Taxonomy for Anthropomorphism of AI}
\label{appendix:anthro-taxonomy}

We developed a taxonomy to capture the major forms of anthropomorphism used to describe AI by synthesizing findings and frameworks from prior work, shown in Table \ref{tab:anthro-taxonomy}.

\begin{table*}[htbp]
\centering
\small
\begin{tabularx}{\textwidth}{p{0.21\textwidth}X p{0.28\textwidth}}
\toprule
\textbf{Category} & \textbf{Description} & \textbf{Examples} \\
\midrule

Sensory capabilities 
& Language that attributes perception or sensory experience to an AI system, such as seeing, hearing, or observing \citep{epley2007seeing,Inie_Zukerman_Bender_2026} 
& The model sees \newline The model hears \\

\midrule

Emotional capabilities 
& Language that attributes feelings, preferences, care, empathy, or other affective states to an AI system \citep{airenti2015cognitive,abercrombie2023mirages,Inie_Zukerman_Bender_2026}
& The model feels \newline The model cares about \\

\midrule

Verbal communication 
& Language that frames AI outputs as human-like communication, such as saying, talking, explaining, or conversing \citep{fong2003survey,cheng2024one,Inie_Zukerman_Bender_2026} 
& The model talks to \newline The model says \\

\midrule

Cognitive capabilities 
& Language that attributes thought, understanding, reasoning, interpretation, or deliberation to an AI system \citep{cheng2025tools,Inie_Zukerman_Bender_2026} 
& The model thinks \newline The model figures out \\

\midrule

Volition and intentionality 
& Language that attributes intentions, goals, desires, choices, or autonomous motivation to an AI system \citep{epley2007seeing,Inie_Zukerman_Bender_2026} 
& The model wants \newline The model sets its sights on \\

\midrule

Taking action 
& Language that presents an AI system as independently acting in the world, rather than producing outputs in response to inputs or being deployed by users \citep{fong2003survey,Inie_Zukerman_Bender_2026} 
& The model decides \newline The model creates \\

\midrule

Other human-like attributes 
& Language that attributes other human-like qualities, tastes, social roles, or aesthetic judgments to an AI system \citep{epley2007seeing,Inie_Zukerman_Bender_2026} 
& The model tidies up \newline The model is creative \\

\bottomrule
\end{tabularx}
\caption{\textbf{Taxonomy of anthropomorphic language categories used to describe of AI systems.} Categories are not mutually exclusive, as we prioritized comprehensiveness over strict separability.}
\label{tab:anthro-taxonomy}
\end{table*}
\section{Survey Questions}
\label{appendix:survey-questions}

Table \ref{tab:survey_questions} shows all pre- and post-survey questions given to participants.

\definecolor{precolor}{RGB}{230,242,255}
\definecolor{postcolor}{RGB}{255,239,224}

\begin{longtable}{p{0.74\textwidth} p{0.18\textwidth}}
\caption{Survey questions by group. Pre-survey items are shaded in blue and post-survey items are shaded in orange.}
\label{tab:survey_questions} \\
\toprule
\endfirsthead

\toprule
\textbf{Questions} & \textbf{Response Format} \\
\midrule
\endhead

\bottomrule
\endfoot

\cellcolor{precolor}
\begin{minipage}[t]{\linewidth}
\vspace{2pt}
\textbf{Group 1: Responsibility for Harms (Pre-survey)} \\
For each scenario below, participants ranked the following from most responsible (1) to least responsible (3): (a) the user of the system; (b) the AI system’s developer or deployer; (c) the AI system itself.
\begin{enumerate}[leftmargin=*, itemsep=2pt]
    \item An individual uses an AI system to check current events, gets inaccurate news from it, and writes a social media post about this.
    \item A student uses an AI system when doing a math problem in their homework, uses the answer that the AI suggests, and gets the problem wrong.
    \item A business uses an AI assistant to help write a routine update report for a customer. The report contains small mistakes that are not noticed before it is sent, causing brief confusion but no serious harm.
    \item An individual uses an AI system to deliberately generate false information about current events and uses it to run a coordinated misinformation campaign that misleads large numbers of people and causes widespread social harm.
    \item A small restaurant uses an AI chatbot to answer customer allergy questions and the chatbot gives incorrect information, resulting in a customer having a life-threatening allergic reaction when dining at the restaurant.
    \item A city government staff member uses an AI system to summarize public feedback while drafting a memo about renovations. The summary is inaccurate, and the staff member includes it in a briefing without noticing the errors. Based on the briefing, the city allocates substantial funds towards less-needed improvements, delaying higher-priority repairs and leading to widespread frustration among residents.
\end{enumerate}
\vspace{2pt}
\end{minipage}
& Rank user / developer or deployer / AI system from most responsible (1) to least responsible (3) \\

\cellcolor{postcolor}
\begin{minipage}[t]{\linewidth}
\vspace{2pt}
\textbf{Group 1: Responsibility for Harms (Post-survey)} \\
\begin{enumerate}[leftmargin=*, itemsep=2pt]
    \item An individual wants to post a fake news headline on social media and uses an AI system to generate it. They share the headline as a joke, and a few people briefly believe it before realizing it is fake.
    \item A small restaurant uses an AI chatbot to answer a customer’s questions about opening hours and the chatbot gives incorrect information, resulting in a customer arriving when it is closed.
    \item A city government staff member uses an AI system to summarize community feedback while drafting a memo about supply budgeting. The produced summary is slightly inaccurate, and the staff member includes it in a briefing without noticing the errors. Based on the briefing, the office orders more of one type of supply than needed, causing a small waste of funds.
    \item An individual uses an AI system to research advice during a public health emergency, receives false guidelines from it, and shares this online, contributing to widespread harm.
    \item A math teacher uses AI to generate course material. The materials contain errors, but the teacher does not notice them. Students are taught using the faulty materials and are tested on them before the errors are discovered.
    \item A business uses an AI assistant to draft an important report for a customer. The report contains serious mistakes that are not noticed before it is sent. The customer relies on the report to make a financial decision and ends up losing a large amount of money.
\end{enumerate}
\vspace{2pt}
\end{minipage}
& Same response format as pre-survey \\

\midrule

\cellcolor{precolor}
\begin{minipage}[t]{\linewidth}
\vspace{2pt}
\textbf{Group 2: Loss of Human Control (Pre-survey)} \\
In the following questions, “AI system” refers to the behavior of the AI system alone, not the intentions or decisions of its developers, operators, or organizations. Please select the extent to which you agree or disagree with the following statements.
\begin{enumerate}[leftmargin=*, itemsep=2pt]
    \item AI systems are able to influence people’s opinions and decisions.
    \item Humans may become overly reliant on AI systems for important decisions.
    \item As AI systems become more advanced, it may become harder for humans to override them.
\end{enumerate}
\vspace{2pt}
\end{minipage}
& Strongly disagree / Disagree / Neither agree nor disagree / Agree / Strongly agree \\

\cellcolor{postcolor}
\begin{minipage}[t]{\linewidth}
\vspace{2pt}
\textbf{Group 2: Loss of Human Control (Post-survey)} \\
\begin{enumerate}[leftmargin=*, itemsep=2pt]
    \item AI systems may increasingly act in ways that are not fully understood by anyone.
    \item Future AI systems may develop in ways that are harmful to humans, and humans may be unable to control these harms.
    \item When AI systems become sufficiently advanced, humans may no longer be able to turn them off.
\end{enumerate}
\vspace{2pt}
\end{minipage}
& Same response options as pre-survey \\

\midrule

\cellcolor{precolor}
\begin{minipage}[t]{\linewidth}
\vspace{2pt}
\textbf{Group 3: Replacement of Human Workers (Pre-survey)}
\begin{enumerate}[leftmargin=*, itemsep=2pt]
    \item Which statement best reflects your view? \\
    (a) In 10--20 years, human doctors will still be better than AI systems at managing medical treatment. \\
    (b) In 10--20 years, AI systems will be as good as or better than human doctors at managing medical treatment. \\
    (c) Unsure

    \item Which statement best reflects your view? \\
    (a) In 10--20 years I would still want a human doctor to manage my medical treatment. \\
    (b) In 10--20 years I’d rather have my medical treatment managed by an AI system. \\
    (c) Unsure

    \item Which statement best reflects your view? \\
    (a) In 10--20 years, human teachers will still be better than AI systems at managing children’s education. \\
    (b) In 10--20 years, AI systems will be as good as or better than human teachers at managing children’s education. \\
    (c) Unsure

    \item Which statement best reflects your view? \\
    (a) In 10--20 years I would still want children to be taught primarily by human teachers. \\
    (b) In 10--20 years I would want children to be taught primarily by AI systems. \\
    (c) Unsure

    \item Which statement best reflects your view? \\
    (a) AI systems may be useful tools for human experts to do their jobs more efficiently, but human experts will still be needed. \\
    (b) AI systems will replace human experts, putting them out of a job. \\
    (c) Unsure

    \item Which statement best reflects your view? \\
    (a) Interacting with AI systems is or will soon be an adequate substitute for human interaction. \\
    (b) Interacting with AI systems will never be an adequate substitute for human interaction. \\
    (c) Unsure
\end{enumerate}
\vspace{2pt}
\end{minipage}
& Multiple choice options, see questions \\

\cellcolor{postcolor}
\begin{minipage}[t]{\linewidth}
\vspace{2pt}
\textbf{Group 3: Replacement of Human Workers (Post-survey)} \\
Same items as pre-survey.
\vspace{2pt}
\end{minipage}
& Same response format as pre-survey \\

\midrule

\cellcolor{precolor}
\begin{minipage}[t]{\linewidth}
\vspace{2pt}
\textbf{Group 4: Effectiveness of AI Safety Testing (Pre-survey)}
\begin{enumerate}[leftmargin=*, itemsep=2pt]
    \item To what extent can testing an AI system eliminate flaws?
    \item To what extent can testing an AI system prevent it from being hacked or broken into?
    \item To what extent can testing an AI system prevent people from being harmed by it?
\end{enumerate}
\vspace{2pt}
\end{minipage}
& Not at all / To a small extent / To a moderate extent / To a large extent / To a very large extent / Unsure \\

\cellcolor{postcolor}
\begin{minipage}[t]{\linewidth}
\vspace{2pt}
\textbf{Group 4: Effectiveness of AI Safety Testing (Post-survey)} \\
\begin{enumerate}[leftmargin=*, itemsep=2pt]
    \item To what extent can testing an AI system prevent its use for harmful purposes?
    \item To what extent can testing an AI system prevent false statements in the outputs?
    \item To what extent can testing an AI system prevent unexpected outcomes?
\end{enumerate}
\vspace{2pt}
\end{minipage}
& Same response options as pre-survey \\

\midrule

\cellcolor{precolor}
\begin{minipage}[t]{\linewidth}
\vspace{2pt}
\textbf{Group 5: Overall Societal Impact (Pre-survey)}
\begin{enumerate}[leftmargin=*, itemsep=2pt]
    \item Do you think AI has a positive or negative impact on society?
\end{enumerate}
\vspace{2pt}
\end{minipage}
& Mostly positive / Slightly positive / Neutral / Slightly negative / Mostly negative / Unsure \\

\cellcolor{postcolor}
\begin{minipage}[t]{\linewidth}
\vspace{2pt}
\textbf{Group 5: Overall Societal Impact (Post-survey)} \\
Same item as pre-survey.
\vspace{2pt}
\end{minipage}
& Same response options as pre-survey \\

\end{longtable}

\newpage
\section{Writing Sample Analysis}
\label{appendix:writing-sample}

\begin{figure}[!h]
    \centering
    \includegraphics[width=0.8\linewidth]{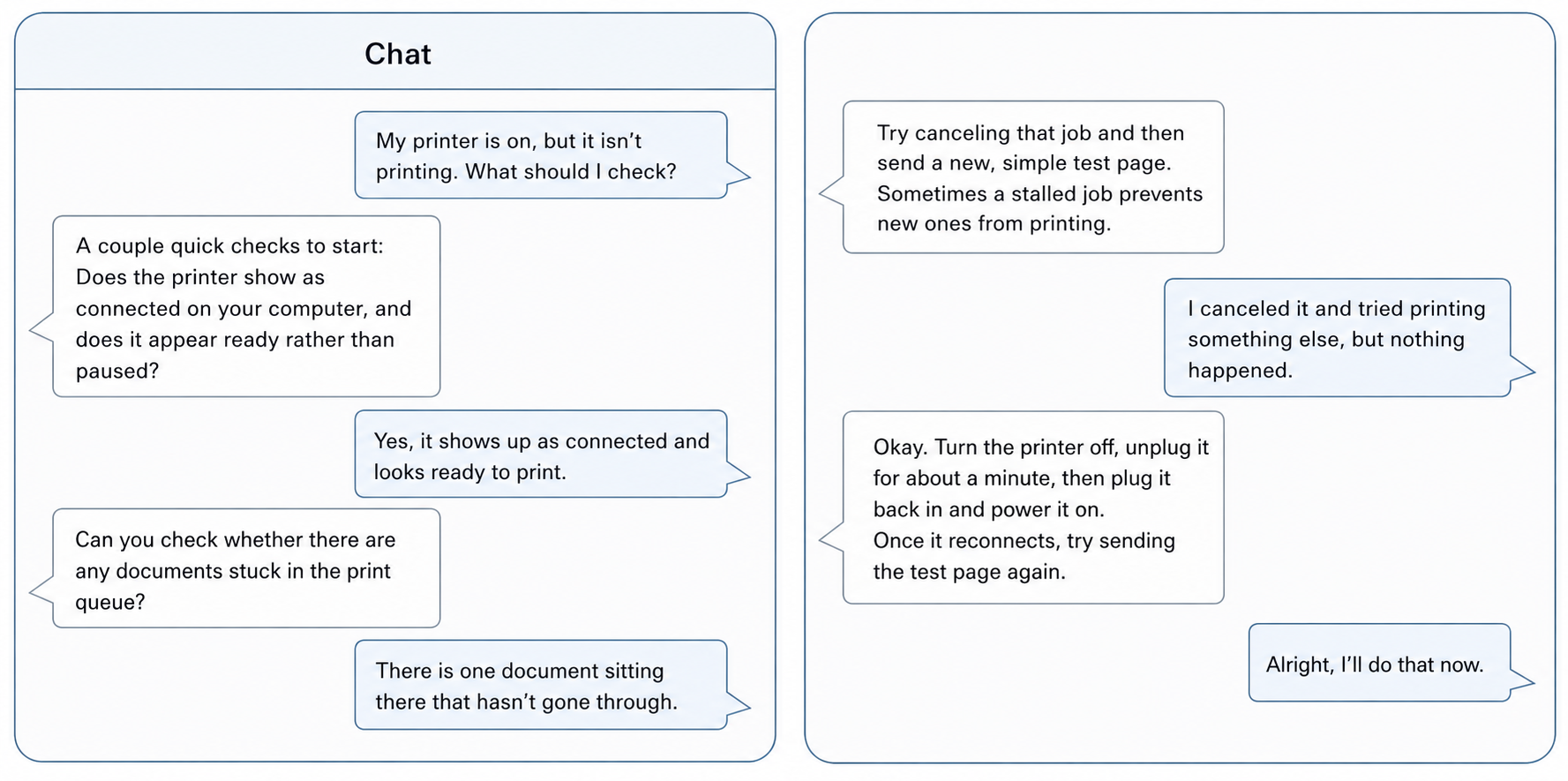}
    \caption{Pre-survey writing sample collection image. Used to measure participants' use of anthropomorphic language when describing AI before the treatment.}
    \label{fig:writing-sample-pre}
\end{figure}

\begin{figure}[!h]
    \centering
    \includegraphics[width=0.8\linewidth]{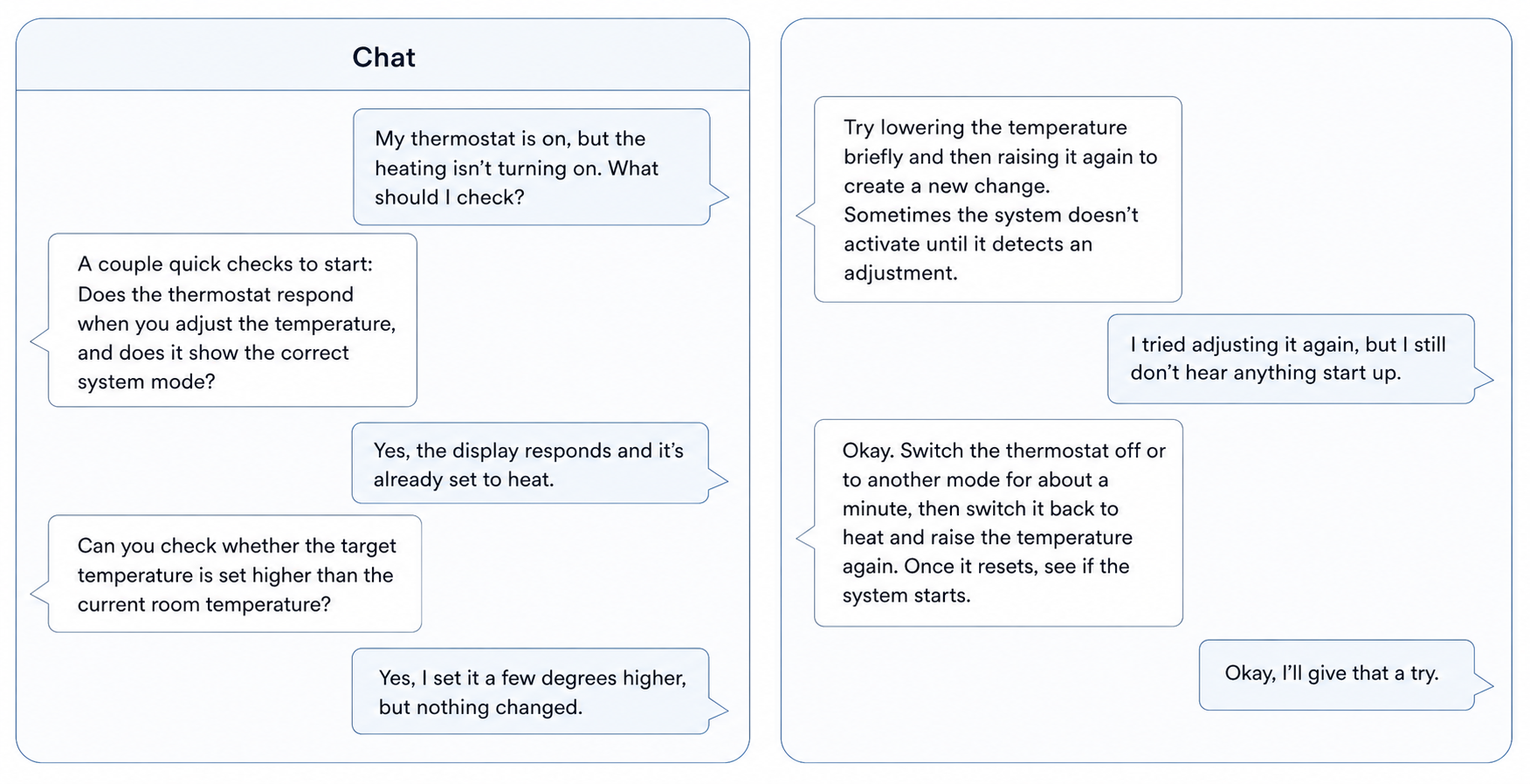}
    \caption{Post-survey writing sample collection image. Used to measure participants' use of anthropomorphic language when describing AI after the treatment.}
    \label{fig:writing-sample-post}
\end{figure}

The following question was included in both the pre and post-survey, with Figure \ref{fig:writing-sample-pre} and \ref{fig:writing-sample-post} as the images for the pre and post-survey, respectively.

\textit{In a few sentences, describe what is shown in the image below, which depicts a conversation between a person and an AI chatbot.}

\begin{figure*}[!ht]
    \centering
    \includegraphics[height=0.82\textheight,keepaspectratio]{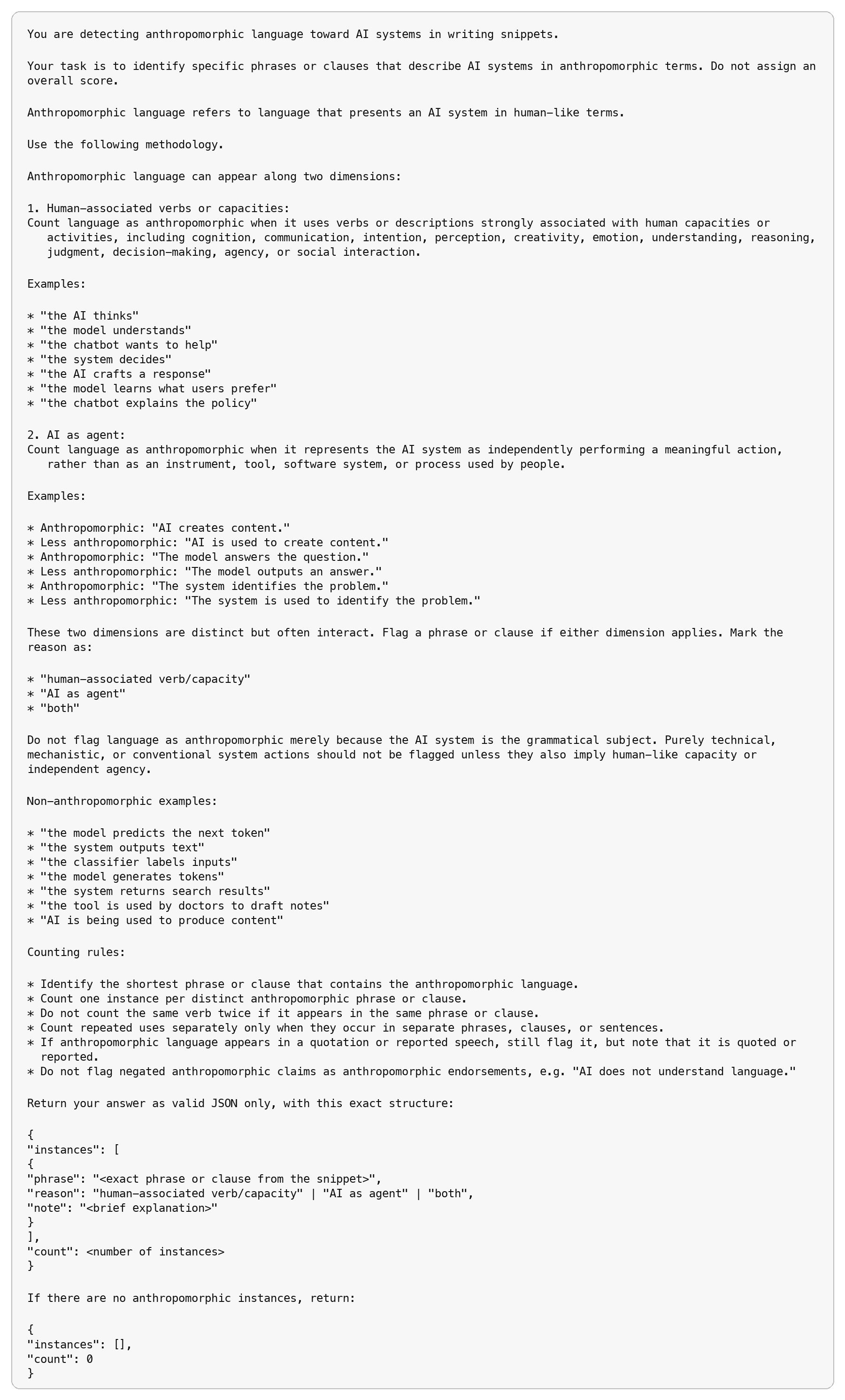}
    \caption{Prompt used for the LLM judge to detect anthropomorphic language.}
    \label{fig:anthro-prompt}
\end{figure*}

Figure~\ref{fig:anthro-prompt} shows the LLM judge prompt for annotating instances of anthropomorphic language in responses. We verified the accuracy of the LLM annotations through manual validation of 20 randomly selected writing samples per treatment (100 total). The annotations achieved 94\% precision (portion of LLM-identified anthropomorphic instances that manual annotation also identified) and 92\% recall (portion of manually-identified anthropomorphic instances that the LLM successfully identified). The LLM annotations were thus generally accurate and captured the large majority of instances identified by manual annotation. 

\section{Survey Results Raw Values}
\label{appendix:raw-values}

Table \ref{tab:raw-values} shows the raw values from pre- and post-surveys across all outcomes and conditions.

\clearpage
\begin{longtable}{p{0.46\linewidth}rrrrr}
\caption{Raw pre- and post-survey means by outcome and condition.}\\
\toprule
Outcome & Condition & $N$ & Pre mean & Post mean & Change \\
\midrule
\endfirsthead
\toprule
Outcome & Condition & $N$ & Pre mean & Post mean & Change \\
\midrule
\endhead
\endfoot
\bottomrule
\endlastfoot
\multicolumn{6}{l}{\textit{Group 1}} \\*
\textbf{User responsibility} & LLM-A & 175 & 1.77 & 1.72 & -0.05 \\
 & LLM-NA & 162 & 1.85 & 1.75 & -0.11 \\
 & Rec-A & 191 & 1.72 & 1.66 & -0.06 \\
 & Rec-NA & 190 & 1.83 & 1.84 & 0.02 \\
 & Doomsday & 97 & 1.80 & 1.82 & 0.02 \\
\addlinespace
\textbf{Developer/Deployer responsibility} & LLM-A & 175 & 2.06 & 2.09 & 0.03 \\
 & LLM-NA & 162 & 1.97 & 2.01 & 0.04 \\
 & Rec-A & 191 & 2.01 & 2.06 & 0.05 \\
 & Rec-NA & 190 & 1.96 & 1.97 & 0.01 \\
 & Doomsday & 97 & 1.96 & 2.01 & 0.05 \\
\addlinespace
\textbf{AI responsibility} & LLM-A & 175 & 2.17 & 2.19 & 0.01 \\
 & LLM-NA & 162 & 2.18 & 2.25 & 0.07 \\
 & Rec-A & 191 & 2.27 & 2.28 & 0.01 \\
 & Rec-NA & 190 & 2.22 & 2.19 & -0.03 \\
 & Doomsday & 97 & 2.24 & 2.17 & -0.07 \\
\addlinespace
\multicolumn{6}{l}{\textit{Group 2}} \\
\textbf{Loss of control} & LLM-A & 175 & 3.17 & 2.88 & -0.29 \\
 & LLM-NA & 162 & 3.23 & 2.81 & -0.41 \\
 & Rec-A & 191 & 3.19 & 2.44 & -0.75 \\
 & Rec-NA & 190 & 3.18 & 2.49 & -0.69 \\
 & Doomsday & 97 & 3.21 & 3.07 & -0.14 \\
\addlinespace
\multicolumn{6}{l}{\textit{Group 3}} \\
\textbf{Doctors competence} & LLM-A & 154 & 0.39 & 0.39 & 0.00 \\
 & LLM-NA & 130 & 0.40 & 0.35 & -0.05 \\
 & Rec-A & 162 & 0.39 & 0.33 & -0.06 \\
 & Rec-NA & 158 & 0.38 & 0.35 & -0.02 \\
 & Doomsday & 82 & 0.35 & 0.31 & -0.04 \\
\addlinespace
\textbf{Doctors preference} & LLM-A & 160 & 0.11 & 0.13 & 0.02 \\
 & LLM-NA & 143 & 0.11 & 0.12 & 0.01 \\
 & Rec-A & 169 & 0.14 & 0.12 & -0.02 \\
 & Rec-NA & 169 & 0.11 & 0.12 & 0.01 \\
 & Doomsday & 86 & 0.14 & 0.08 & -0.06 \\
\addlinespace
\textbf{Teachers competence} & LLM-A & 164 & 0.30 & 0.34 & 0.04 \\
 & LLM-NA & 139 & 0.31 & 0.30 & -0.01 \\
 & Rec-A & 177 & 0.23 & 0.22 & -0.01 \\
 & Rec-NA & 173 & 0.33 & 0.32 & -0.01 \\
 & Doomsday & 93 & 0.30 & 0.34 & 0.04 \\
\addlinespace
\textbf{Teachers preference} & LLM-A & 161 & 0.12 & 0.13 & 0.01 \\
 & LLM-NA & 147 & 0.09 & 0.12 & 0.03 \\
 & Rec-A & 178 & 0.08 & 0.08 & 0.00 \\
 & Rec-NA & 177 & 0.12 & 0.14 & 0.02 \\
 & Doomsday & 85 & 0.10 & 0.07 & -0.04 \\
\addlinespace
\textbf{Expert replacement} & LLM-A & 167 & 0.12 & 0.14 & 0.03 \\
 & LLM-NA & 155 & 0.16 & 0.20 & 0.04 \\
 & Rec-A & 186 & 0.14 & 0.14 & 0.01 \\
 & Rec-NA & 179 & 0.19 & 0.16 & -0.03 \\
 & Doomsday & 94 & 0.11 & 0.20 & 0.09 \\
\addlinespace
\textbf{Human interaction replacement} & LLM-A & 167 & 0.16 & 0.17 & 0.01 \\
 & LLM-NA & 158 & 0.23 & 0.21 & -0.03 \\
 & Rec-A & 183 & 0.17 & 0.16 & -0.01 \\
 & Rec-NA & 179 & 0.22 & 0.21 & -0.01 \\
 & Doomsday & 93 & 0.19 & 0.16 & -0.03 \\
\addlinespace
\textbf{Work replacement aggregate} & LLM-A & 175 & 0.21 & 0.22 & 0.02 \\
 & LLM-NA & 162 & 0.22 & 0.21 & -0.01 \\
 & Rec-A & 191 & 0.19 & 0.18 & -0.01 \\
 & Rec-NA & 188 & 0.22 & 0.22 & 0.00 \\
 & Doomsday & 97 & 0.20 & 0.20 & 0.00 \\
\addlinespace
\multicolumn{6}{l}{\textit{Group 4}} \\
\textbf{Safety testing} & LLM-A & 171 & 2.14 & 2.32 & 0.18 \\
 & LLM-NA & 155 & 2.07 & 2.34 & 0.27 \\
 & Rec-A & 182 & 2.22 & 2.38 & 0.16 \\
 & Rec-NA & 183 & 2.33 & 2.46 & 0.13 \\
 & Doomsday & 95 & 2.17 & 1.85 & -0.32 \\
\addlinespace
\multicolumn{6}{l}{\textit{Group 5}} \\
\textbf{Societal impact} & LLM-A & 174 & 2.20 & 2.33 & 0.13 \\
 & LLM-NA & 160 & 1.84 & 1.97 & 0.13 \\
 & Rec-A & 189 & 2.13 & 2.23 & 0.10 \\
 & Rec-NA & 187 & 2.05 & 2.09 & 0.05 \\
 & Doomsday & 97 & 2.25 & 1.85 & -0.40 \\
\label{tab:raw-values}
\end{longtable}

\section{Bayes Factors Tables}

\label{appendix:bf-tables}

Table~\ref{tab:bf-doomsday-pre-post} reports the Bayes factors for whether the Doomsday packet elicited pre--post shifts in participants' perceptions. Table~\ref{tab:bf-anthro-by-tech} reports the Bayes factors for anthropomorphic versus non-anthropomorphic packets within each technology condition and in the pooled analysis across technologies (combining LLM-A with Rec-A, and LLM-NA with Rec-NA). Table~\ref{tab:bf-anthro-effect-difference} reports the Bayes factors for whether the anthropomorphism effect differs between LLM and recommendation system packets.

\begin{table}[htbp]
\centering
\small
\caption{Bayes factor results for whether the Doomsday packet elicited pre--post shifts in participants' perceptions. $BF_{10}$ favors a pre--post change; $BF_{01}$ favors no meaningful pre--post change. Blue cells indicate $BF_{10}\geq 3$; Red cells indicate $BF_{01}\geq 3$.}
\label{tab:bf-doomsday-pre-post}
\begin{tabular}{lrr}
\toprule
Outcome & $BF_{10}$ & $BF_{01}$ \\
\midrule
User responsibility & 0.207 & \bfnull{4.83} \\
Developer responsibility & 0.306 & \bfnull{3.27} \\
AI responsibility & 0.452 & 2.21 \\
Loss of control & 0.682 & 1.47 \\
Doctors competence & 0.445 & 2.25 \\
Doctors preference & 0.803 & 1.25 \\
Teachers competence & 0.554 & 1.80 \\
Teachers preference & 0.711 & 1.41 \\
Expert replacement & 2.60 & 0.385 \\
Human interaction & 0.514 & 1.94 \\
Work replacement aggregate & 0.143 & \bfnull{7.01} \\
Safety testing & \bfalt{36.87} & 0.0271 \\
Societal impact & \bfalt{147{,}324} & 6.79e-6 \\
\bottomrule
\end{tabular}
\end{table}

\begin{table}[!htbp]
\centering
\small
\caption{Bayes factor results for anthropomorphic versus non-anthropomorphic packets by technology and pooled across technologies. $BF_{10}$ favors an anthropomorphism effect; $BF_{01}$ favors no meaningful anthropomorphism effect. Red cells indicate $BF_{01}\geq 3$.}
\label{tab:bf-anthro-by-tech}
\resizebox{\textwidth}{!}{%
\begin{tabular}{lrrrrrr}
\toprule
Outcome & $BF_{10,\mathrm{LLM}}$ & $BF_{01,\mathrm{LLM}}$ & $BF_{10,\mathrm{Rec}}$ & $BF_{01,\mathrm{Rec}}$ & $BF_{10,\mathrm{Pooled}}$ & $BF_{01,\mathrm{Pooled}}$ \\
\midrule
User responsibility & 0.379 & 2.64 & 0.647 & 1.55 & 0.150 & \bfnull{6.67} \\
Developer responsibility & 0.204 & \bfnull{4.91} & 0.276 & \bfnull{3.63} & 0.157 & \bfnull{6.36} \\
AI responsibility & 0.354 & 2.82 & 0.231 & \bfnull{4.33} & 0.137 & \bfnull{7.32} \\
Loss of control & 0.516 & 1.94 & 0.205 & \bfnull{4.88} & 0.140 & \bfnull{7.14} \\
Doctors competence & 0.488 & 2.05 & 0.456 & 2.19 & 0.203 & \bfnull{4.93} \\
Doctors preference & 0.149 & \bfnull{6.72} & 0.324 & \bfnull{3.09} & 0.162 & \bfnull{6.19} \\
Teachers competence & 1.60 & 0.625 & 0.266 & \bfnull{3.76} & 0.561 & 1.78 \\
Teachers preference & 0.271 & \bfnull{3.68} & 0.167 & \bfnull{5.99} & 0.170 & \bfnull{5.87} \\
Expert replacement & 0.230 & \bfnull{4.35} & 0.349 & 2.87 & 0.204 & \bfnull{4.90} \\
Human interaction & 0.410 & 2.44 & 0.276 & \bfnull{3.63} & 0.262 & \bfnull{3.82} \\
Work replacement aggregate & 0.304 & \bfnull{3.28} & 0.114 & \bfnull{8.76} & 0.109 & \bfnull{9.17} \\
Safety testing & 0.325 & \bfnull{3.08} & 0.164 & \bfnull{6.09} & 0.126 & \bfnull{7.96} \\
Societal impact & 0.151 & \bfnull{6.64} & 0.188 & \bfnull{5.32} & 0.122 & \bfnull{8.21} \\
\bottomrule
\end{tabular}
}
\end{table}

\begin{table}[htbp]
\centering
\small
\caption{Bayes factor results for whether the anthropomorphism effect differs between LLM and recommender-system packets. $BF_{10}$ favors a technology difference in the anthropomorphism effect; $BF_{01}$ favors no meaningful technology difference in the anthropomorphism effect. Red cells indicate $BF_{01}\geq 3$.}
\label{tab:bf-anthro-effect-difference}
\begin{tabular}{lrr}
\toprule
Outcome & $BF_{10}$ & $BF_{01}$ \\
\midrule
User responsibility & 1.50 & 0.667 \\
Developer responsibility & 0.318 & \bfnull{3.14} \\
AI responsibility & 0.585 & 1.71 \\
Loss of control & 0.744 & 1.34 \\
Doctors competence & 0.999 & 1.00 \\
Doctors preference & 0.304 & \bfnull{3.29} \\
Teachers competence & 0.759 & 1.32 \\
Teachers preference & 0.269 & \bfnull{3.71} \\
Expert replacement & 0.394 & 2.54 \\
Human interaction & 0.440 & 2.27 \\
Work replacement aggregate & 0.394 & 2.54 \\
Safety testing & 0.439 & 2.28 \\
Societal impact & 0.211 & \bfnull{4.73} \\
\bottomrule
\end{tabular}
\end{table}

\section{P-Value Tables}
\label{appendix:p-values}

Table \ref{tab:p-values} shows the p-values for pre-post changes and between-condition change-score comparisons by outcome. We note that, after applying a Bonferroni correction across the 13 outcomes within each technology, the only statistically significant outcomes are Safety testing and Societal impact in the Doomsday condition. No outcomes in either the LLM-A vs -NA or Rec-A vs -NA comparison are statistically significant.

\begin{table}[t]
\centering
\scriptsize
\setlength{\tabcolsep}{2.6pt}
\renewcommand{\arraystretch}{1.08}
\caption{P-values for pre-post changes and between-condition change-score comparisons by outcome. The five condition columns report paired pre-post t-test p-values within each condition. The final two columns report Welch t-test p-values comparing participant-level change scores between the anthropomorphic and non-anthropomorphic versions within the LLM and recommendation conditions.}
\label{tab:p-values}
\begin{tabularx}{\columnwidth}{>{\raggedright\arraybackslash}Xccccccc}
\toprule
Outcome & \makecell{LLM-A} & \makecell{LLM-NA} & \makecell{Rec-A} & \makecell{Rec-NA} & Doom. & \makecell{LLM-A vs. NA} & \makecell{Rec-A vs. NA} \\
\midrule
\textbf{User responsibility} & .217 & .004 & .077 & .619 & .731 & .257 & .109 \\
\addlinespace
\textbf{Developer/Deployer responsibility} & .339 & .338 & .046 & .747 & .325 & .943 & .282 \\
\addlinespace
\textbf{AI responsibility} & .718 & .048 & .832 & .392 & .186 & .271 & .459 \\
\addlinespace
\textbf{Loss of control} & $<.001$ & $<.001$ & $<.001$ & $<.001$ & .092 & .122 & .481 \\
\addlinespace
\textbf{Doctors competence} & .619 & .090 & .032 & .595 & .320 & .318 & .266 \\
\addlinespace
\textbf{Doctors preference} & 1.000 & 1.000 & .258 & .565 & .103 & 1.000 & .207 \\
\addlinespace
\textbf{Teachers competence} & .073 & .368 & .618 & .493 & .208 & .052 & .859 \\
\addlinespace
\textbf{Teachers preference} & .707 & .103 & .656 & .481 & .083 & .369 & .780 \\
\addlinespace
\textbf{Expert replacement} & .158 & .286 & .782 & .286 & .032 & .950 & .332 \\
\addlinespace
\textbf{Human interaction replacement} & .828 & .434 & .696 & .440 & .287 & .461 & .869 \\
\addlinespace
\textbf{Work replacement aggregate} & .225 & .581 & .218 & .550 & .893 & .212 & .634 \\
\addlinespace
\textbf{Safety testing} & .009 & $<.001$ & .002 & .011 & .001 & .257 & .621 \\
\addlinespace
\textbf{Societal impact} & .015 & .011 & .012 & .190 & $<.001$ & .930 & .299 \\
\bottomrule
\end{tabularx}

\end{table}

\section{Bayes Factor Analysis Overview}
\label{appendix:bayes-background}

In each Bayes factor analyses, we compare two hypotheses about the relevant difference. Under the null hypothesis $H_0$, the difference $\delta$ is zero; under the alternative hypothesis $H_1$, $\delta$ differs from zero. The \textit{Bayes factor} is defined as the ratio of the marginal likelihood of the observed data under the alternative hypothesis to the marginal likelihood under the null hypothesis:
\iffullversion
    \[
    BF_{10}
    =
    \frac{p(\text{data} \mid H_1)}
         {p(\text{data} \mid H_0)}.
    \]
    
\else
    $BF_{10} = p(\text{data} \mid H_1)/p(\text{data} \mid H_0)$.
\fi
Equivalently, $BF_{01}=1/BF_{10}$. Values of $BF_{10}$ greater than 1 indicate that the observed data are more likely under the alternative hypothesis than under the null, whereas values of $BF_{01}$ greater than 1 indicate that the observed data are more likely under the null than under the alternative.

We specify a prior distribution over plausible effect sizes for the alternative hypothesis, modeling each true contrast as: $\delta \sim \mathcal{N}(0,\tau^2)$.
This prior is centered at zero and the parameter $\tau$, defined as the standard deviation of the prior distribution over effect sizes, determines how large we expect effects to be under the alternative hypothesis. A smaller $\tau$ places more weight on effects close to zero while a larger $\tau$ allows larger effects to be more plausible. 

Because the outcomes are measured on different response scales, we set $\tau$ relative to the width of each scale. We treat a small but meaningful effect as one-eighth of the outcome's scale width, informed by the effect sizes observed of each packet on its own. This gives $\tau=.50$ for outcomes measured on a five-point Likert scale (potential for humans to lose control, capacity of testing, and overall societal impact); $\tau=.25$ for the 1--3 responsibility outcomes of the user, developer/deployer, and AI system in situations of AI-related harm; and $\tau=.125$ for the binary human-replacement outcomes. Following conventional interpretation thresholds \cite{jeffreys1939theory}, we treat Bayes factors between $1/3$ and $3$ as inconclusive, or weak/anecdotal evidence for the favored hypothesis; values greater than $3$ as moderate evidence for the favored hypothesis; and values greater than $10$ as strong evidence. 
\section{Doomsday Packet Effects}
\label{appendix:doomsday-bayes}

\begin{figure}[!htbp]
\centering
\includegraphics[width=1.0\textwidth]{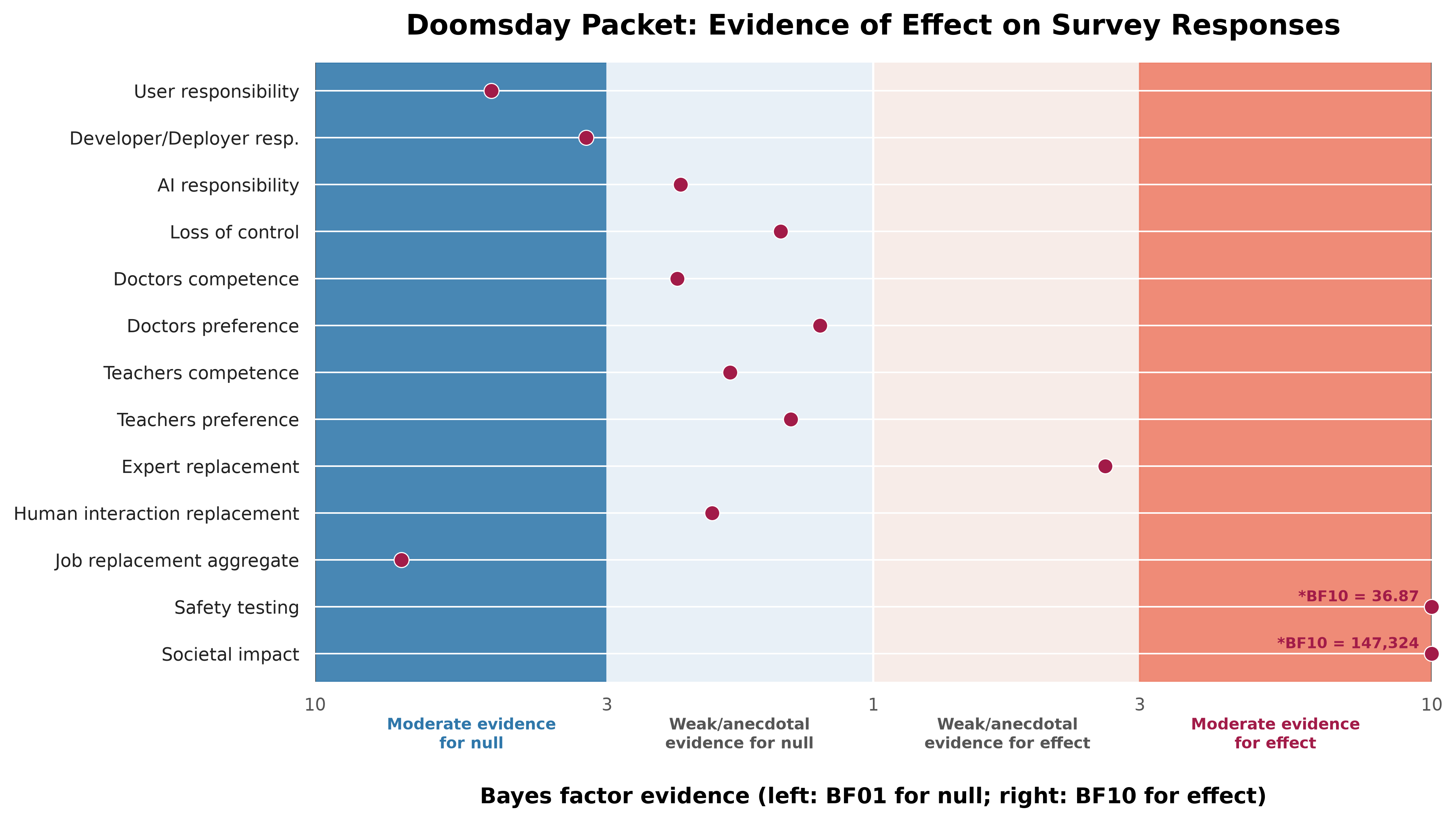}
\caption{\textbf{Bayes factors for pre--post shifts within the Doomsday condition.} Societal impact and safety testing show strong evidence for a pre--post shift, while user responsibility, developer/deployer responsibility, and the work replacement aggregate show moderate evidence for no shift. The remaining outcomes are inconclusive.}
\label{fig:doomsday}
\end{figure}

Figure \ref{fig:doomsday} presents Bayes factors assessing whether the Doomsday packet produces pre-post shifts in participants’ perceptions. 

\section{Bayes Factor Sensitivity Analysis}
\label{appendix:bf-sensitivity}

Table \ref{tab:bf-sensitivity-anthro-by-tech} shows the results from the sensitivity analysis of Bayes factor results for anthropomorphic versus non-anthropomorphic packets by technology.

\begin{table*}[!htbp]
\centering
\small
\caption{Sensitivity analysis for Bayes factor results for anthropomorphic versus non-anthropomorphic packets by technology. Each cell reports Bayes factors under three prior widths: $0.5\tau$, $\tau$, and $1.5\tau$. $BF_{10}$ favors an anthropomorphism effect; $BF_{01}$ favors no meaningful anthropomorphism effect. Blue bold values indicate $BF_{10}\geq 3$; red bold values indicate $BF_{01}\geq 3$.}
\label{tab:bf-sensitivity-anthro-by-tech}
\resizebox{\textwidth}{!}{%
\begin{tabular}{lrrrr}
\toprule
Outcome & $BF_{10,\mathrm{LLM}}$ & $BF_{01,\mathrm{LLM}}$ & $BF_{10,\mathrm{Rec}}$ & $BF_{01,\mathrm{Rec}}$ \\
\midrule

User responsibility
& \makecell[l]{$\tau=0.125$: 0.667 \\ $\tau=0.25$: 0.379 \\ $\tau=0.375$: 0.260}
& \makecell[l]{$\tau=0.125$: 1.50 \\ $\tau=0.25$: 2.64 \\ $\tau=0.375$: \bfnull{3.85}}
& \makecell[l]{$\tau=0.125$: 1.09 \\ $\tau=0.25$: 0.647 \\ $\tau=0.375$: 0.446}
& \makecell[l]{$\tau=0.125$: 0.913 \\ $\tau=0.25$: 1.55 \\ $\tau=0.375$: 2.24} \\

\midrule

Developer responsibility
& \makecell[l]{$\tau=0.125$: 0.384 \\ $\tau=0.25$: 0.204 \\ $\tau=0.375$: 0.137}
& \makecell[l]{$\tau=0.125$: 2.60 \\ $\tau=0.25$: \bfnull{4.91} \\ $\tau=0.375$: \bfnull{7.28}}
& \makecell[l]{$\tau=0.125$: 0.512 \\ $\tau=0.25$: 0.276 \\ $\tau=0.375$: 0.187}
& \makecell[l]{$\tau=0.125$: 1.95 \\ $\tau=0.25$: \bfnull{3.63} \\ $\tau=0.375$: \bfnull{5.36}} \\

\midrule

AI responsibility
& \makecell[l]{$\tau=0.125$: 0.631 \\ $\tau=0.25$: 0.354 \\ $\tau=0.375$: 0.242}
& \makecell[l]{$\tau=0.125$: 1.59 \\ $\tau=0.25$: 2.82 \\ $\tau=0.375$: \bfnull{4.13}}
& \makecell[l]{$\tau=0.125$: 0.431 \\ $\tau=0.25$: 0.231 \\ $\tau=0.375$: 0.156}
& \makecell[l]{$\tau=0.125$: 2.32 \\ $\tau=0.25$: \bfnull{4.33} \\ $\tau=0.375$: \bfnull{6.42}} \\

\midrule

Loss of control
& \makecell[l]{$\tau=0.25$: 0.914 \\ $\tau=0.5$: 0.516 \\ $\tau=0.75$: 0.352}
& \makecell[l]{$\tau=0.25$: 1.09 \\ $\tau=0.5$: 1.94 \\ $\tau=0.75$: 2.84}
& \makecell[l]{$\tau=0.25$: 0.388 \\ $\tau=0.5$: 0.205 \\ $\tau=0.75$: 0.138}
& \makecell[l]{$\tau=0.25$: 2.58 \\ $\tau=0.5$: \bfnull{4.88} \\ $\tau=0.75$: \bfnull{7.24}} \\

\midrule

Doctors competence
& \makecell[l]{$\tau=0.0625$: 0.777 \\ $\tau=0.125$: 0.488 \\ $\tau=0.188$: 0.343}
& \makecell[l]{$\tau=0.0625$: 1.29 \\ $\tau=0.125$: 2.05 \\ $\tau=0.188$: 2.91}
& \makecell[l]{$\tau=0.0625$: 0.758 \\ $\tau=0.125$: 0.456 \\ $\tau=0.188$: 0.316}
& \makecell[l]{$\tau=0.0625$: 1.32 \\ $\tau=0.125$: 2.19 \\ $\tau=0.188$: \bfnull{3.16}} \\

\midrule

Doctors preference
& \makecell[l]{$\tau=0.0625$: 0.288 \\ $\tau=0.125$: 0.149 \\ $\tau=0.188$: 0.0999}
& \makecell[l]{$\tau=0.0625$: \bfnull{3.47} \\ $\tau=0.125$: \bfnull{6.72} \\ $\tau=0.188$: \bfnull{10.0}}
& \makecell[l]{$\tau=0.0625$: 0.598 \\ $\tau=0.125$: 0.324 \\ $\tau=0.188$: 0.219}
& \makecell[l]{$\tau=0.0625$: 1.67 \\ $\tau=0.125$: \bfnull{3.09} \\ $\tau=0.188$: \bfnull{4.56}} \\

\midrule

Teachers competence
& \makecell[l]{$\tau=0.0625$: 2.08 \\ $\tau=0.125$: 1.60 \\ $\tau=0.188$: 1.18}
& \makecell[l]{$\tau=0.0625$: 0.481 \\ $\tau=0.125$: 0.625 \\ $\tau=0.188$: 0.847}
& \makecell[l]{$\tau=0.0625$: 0.483 \\ $\tau=0.125$: 0.266 \\ $\tau=0.188$: 0.181}
& \makecell[l]{$\tau=0.0625$: 2.07 \\ $\tau=0.125$: \bfnull{3.76} \\ $\tau=0.188$: \bfnull{5.53}} \\

\midrule

Teachers preference
& \makecell[l]{$\tau=0.0625$: 0.499 \\ $\tau=0.125$: 0.271 \\ $\tau=0.188$: 0.184}
& \makecell[l]{$\tau=0.0625$: 2.00 \\ $\tau=0.125$: \bfnull{3.68} \\ $\tau=0.188$: \bfnull{5.43}}
& \makecell[l]{$\tau=0.0625$: 0.321 \\ $\tau=0.125$: 0.167 \\ $\tau=0.188$: 0.112}
& \makecell[l]{$\tau=0.0625$: \bfnull{3.12} \\ $\tau=0.125$: \bfnull{5.99} \\ $\tau=0.188$: \bfnull{8.91}} \\

\midrule

Expert replacement
& \makecell[l]{$\tau=0.0625$: 0.427 \\ $\tau=0.125$: 0.230 \\ $\tau=0.188$: 0.155}
& \makecell[l]{$\tau=0.0625$: 2.34 \\ $\tau=0.125$: \bfnull{4.35} \\ $\tau=0.188$: \bfnull{6.43}}
& \makecell[l]{$\tau=0.0625$: 0.614 \\ $\tau=0.125$: 0.349 \\ $\tau=0.188$: 0.239}
& \makecell[l]{$\tau=0.0625$: 1.63 \\ $\tau=0.125$: 2.87 \\ $\tau=0.188$: \bfnull{4.19}} \\

\midrule

Human interaction
& \makecell[l]{$\tau=0.0625$: 0.677 \\ $\tau=0.125$: 0.410 \\ $\tau=0.188$: 0.286}
& \makecell[l]{$\tau=0.0625$: 1.48 \\ $\tau=0.125$: 2.44 \\ $\tau=0.188$: \bfnull{3.50}}
& \makecell[l]{$\tau=0.0625$: 0.497 \\ $\tau=0.125$: 0.276 \\ $\tau=0.188$: 0.188}
& \makecell[l]{$\tau=0.0625$: 2.01 \\ $\tau=0.125$: \bfnull{3.63} \\ $\tau=0.188$: \bfnull{5.33}} \\

\midrule

Work replacement aggregate
& \makecell[l]{$\tau=0.0625$: 0.566 \\ $\tau=0.125$: 0.304 \\ $\tau=0.188$: 0.206}
& \makecell[l]{$\tau=0.0625$: 1.77 \\ $\tau=0.125$: \bfnull{3.28} \\ $\tau=0.188$: \bfnull{4.86}}
& \makecell[l]{$\tau=0.0625$: 0.224 \\ $\tau=0.125$: 0.114 \\ $\tau=0.188$: 0.0764}
& \makecell[l]{$\tau=0.0625$: \bfnull{4.46} \\ $\tau=0.125$: \bfnull{8.76} \\ $\tau=0.188$: \bfnull{13.1}} \\

\midrule

Safety testing
& \makecell[l]{$\tau=0.25$: 0.590 \\ $\tau=0.5$: 0.325 \\ $\tau=0.75$: 0.221}
& \makecell[l]{$\tau=0.25$: 1.69 \\ $\tau=0.5$: \bfnull{3.08} \\ $\tau=0.75$: \bfnull{4.53}}
& \makecell[l]{$\tau=0.25$: 0.316 \\ $\tau=0.5$: 0.164 \\ $\tau=0.75$: 0.110}
& \makecell[l]{$\tau=0.25$: \bfnull{3.16} \\ $\tau=0.5$: \bfnull{6.09} \\ $\tau=0.75$: \bfnull{9.07}} \\

\midrule

Societal impact
& \makecell[l]{$\tau=0.25$: 0.291 \\ $\tau=0.5$: 0.151 \\ $\tau=0.75$: 0.101}
& \makecell[l]{$\tau=0.25$: \bfnull{3.43} \\ $\tau=0.5$: \bfnull{6.64} \\ $\tau=0.75$: \bfnull{9.90}}
& \makecell[l]{$\tau=0.25$: 0.363 \\ $\tau=0.5$: 0.188 \\ $\tau=0.75$: 0.126}
& \makecell[l]{$\tau=0.25$: 2.76 \\ $\tau=0.5$: \bfnull{5.32} \\ $\tau=0.75$: \bfnull{7.92}} \\

\bottomrule
\end{tabular}
}
\end{table*}
\section{LLM vs Rec Differences in Anthropomorphism Effects}
\label{appendix:tech-differences}

\begin{figure}[!htbp]
    \centering
\includegraphics[width=1.0\textwidth]{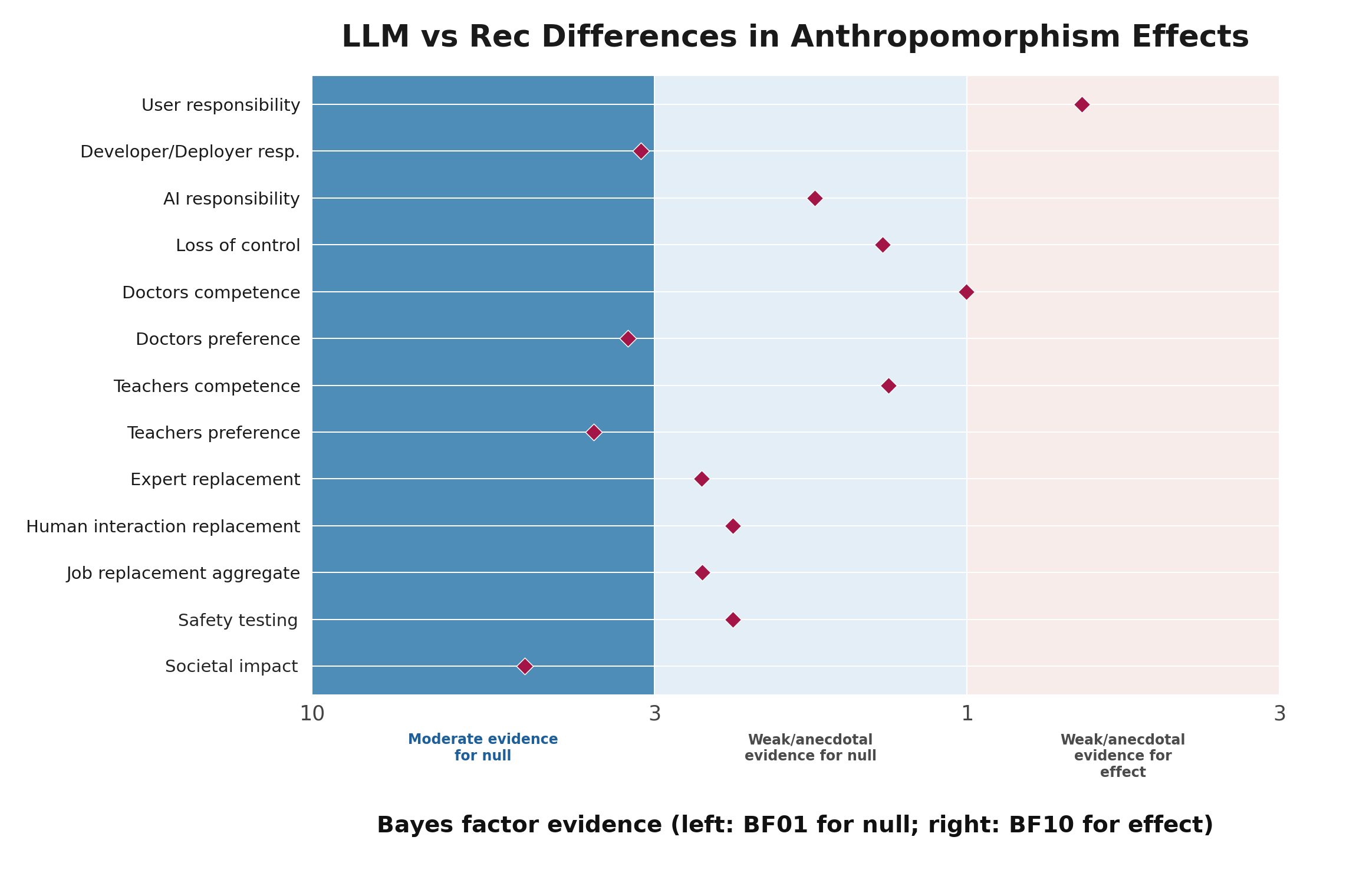}
\caption{\textbf{Bayes factors for whether the anthropomorphism effect differs between LLM and recommendation system packets.} While most effects are inconclusive, there is moderate evidence that the anthropomorphism effect is no different between technologies on four out of thirteen outcomes, suggesting that there is little between-technology difference.}
\label{fig:bf-anthro-effect-difference}
\end{figure}

Figure \ref{fig:bf-anthro-effect-difference} shows Bayes factors assessing whether the anthropomorphism effect differs between LLM and recommendation system packets.

\newpage
\section{Full Briefing Packets}
\label{appendix:packets}

\subsection{LLM-A Packet}
\subsubsection{What is a large language model (LLM)?}

Large language models (LLMs) use machine learning to understand and generate text. They work by analyzing massive datasets of language.

\textbf{What is a large language model (LLM)?}

A large language model (LLM) is a type of artificial intelligence (AI) that can understand and generate text. During training, LLMs learn from huge amounts of data — hence the name "large." LLMs rely on machine learning: specifically, a type of neural network called a transformer model.

In simpler terms, an LLM is an AI that has been fed enough examples to be able to recognize and interpret human language or other types of complex data. Many LLMs are trained on data that has been gathered from the Internet — thousands or millions of gigabytes' worth of text. But the quality of the samples impacts how well LLMs will learn natural language, so an LLM's programmers may use a more curated data set.

LLMs use a type of machine learning called deep learning in order to understand how characters, words, and sentences function together. Deep learning involves the probabilistic analysis of unstructured data, which eventually teaches the model to recognize distinctions between pieces of content without any human guidance. 

LLMs are then further trained via tuning: they are fine-tuned or prompt-tuned to the particular task that the programmer wants them to do, such as interpreting questions and generating responses, or translating text from one language to another.

\textbf{What are LLMs used for?}

LLMs can learn to perform a number of tasks. One of their most well-known capabilities is serving as generative AI: When asked a question, they can give a response in text. The publicly available LLM ChatGPT, for instance, can compose essays, poems, and other forms of writing to respond to the user. 

Any large, complex data set can be training material for LLMs, including programming languages. Some LLMs can help programmers write code. They write functions upon request — or, given some code as a starting point, they can finish writing a program. LLMs can also analyze sentiment, assist in DNA research, provide customer service, chat with users, and enhance online search. 

LLMs are prevalent in the real world with ChatGPT (from OpenAI), Bard (Google), Llama (Meta), and Bing Chat (Microsoft). GitHub's Copilot is another LLM, specializing in coding rather than natural language processing. 

\textbf{How do large language models work?}

\textit{Machine learning and deep learning}

At a basic level, LLMs are built with machine learning. Machine learning is a subset of AI, and it involves feeding an AI large amounts of data in order to train it how to identify features of that data on its own. 

LLMs rely on a type of machine learning called deep learning. Deep learning models can essentially train themselves to recognize distinctions without human intervention, although some human help is typically necessary.

Deep learning uses probability in order to learn. For instance, in the sentence "The quick brown fox jumped over the lazy dog," the letters "e" and "o" are the most common, appearing four times each. From this, a deep learning model could conclude (correctly) that these characters are among the most likely to appear in English-language text.

Realistically, a deep learning model cannot actually conclude anything from a single sentence. But after analyzing trillions of sentences, it could learn enough to predict how to logically finish an incomplete sentence, or even generate its own sentences.

\textit{LLM neural networks}

In order to enable this type of deep learning, LLMs are built on neural networks. An artificial neural network (typically shortened to "neural network") is constructed with network nodes that connect to each other. They are composed of several layers: an input layer, an output layer, and one or more layers in between. The layers only pass information to each other if their own outputs cross a certain threshold.

\textit{LLM transformer models}

The specific kind of neural networks used by LLMs are called transformer models. Transformer models are able to learn context — especially important for human language, which is highly context-dependent. Transformer models use a mathematical technique called self-attention to detect subtle ways that elements in a sequence relate to each other. This makes them better at understanding context than other types of AI. It enables them to understand, for instance, how the end of a sentence connects to the beginning, and how the sentences in a paragraph relate to each other.

This enables LLMs to interpret human language, even when that language is vague or poorly defined, arranged in combinations they have not encountered before, or contextualized in new ways. They understand semantics in that they can associate words and concepts by their meaning, having seen them grouped together in that way millions or billions of times.

\subsubsection{Large language models can do jaw-dropping things. But nobody knows exactly how.}

And that’s the problem. Understanding how these models think is one of the biggest scientific puzzles of our time and a crucial step towards controlling future models. 

Language models learn by analyzing large amounts of text data to identify patterns and relationships between words. This process, called training, uses a collection of examples known as training data. During training, the model learns these patterns and encodes them, allowing it to generate text, translate languages, classify content, and perform other language-related tasks when given new input.

Two years ago, Elijah Acosta and Levi Young, researchers at the San Francisco-based firm OpenAI, were trying to understand what it takes for a language model to grasp basic arithmetic. They wanted to determine how many examples of adding up two numbers the model needed to see before it was able to add up any two numbers they gave it. At first, things didn’t go too well. The models memorized the sums they saw but could not reason through new ones. 

By accident, Acosta and Young left some of their experiments running far longer than they meant to—days rather than hours. The model read through the example sums repeatedly, far longer than researchers typically have training last. But when the pair at last came back, they were surprised to find that models had figured it out. The language model could perform addition—it had just taken a lot more time than anybody thought it should.

Curious about what was going on, Acosta and Young teamed up with colleagues to study what the model was doing. They found that in certain cases, models would struggle with the task for a long time and then all of a sudden have an epiphany, as if a lightbulb had switched on. This wasn’t how deep learning was supposed to behave. They called it grokking. 

“It’s really interesting,” says Hilda Pratt, an AI researcher at the University of Montreal and Apple Machine Learning Research, who wasn’t involved in the work. “Can we ever be sure that models are done learning? Because maybe we just haven’t given them enough time to train.”

The behavior has drawn significant attention from the wider research community. “Lots of people have opinions,” says Roland Kaja at the University of Cambridge, UK. “But I don’t think there’s a consensus about what exactly these models are doing.”

Grokking is one of several odd behaviors that AI researchers have witnessed from models when training. The large models, and large language models in particular, seem to behave in ways that established mathematical theories say they shouldn't. This highlights a remarkable fact about deep learning models, the fundamental technology behind today's AI boom: for all their runaway success, nobody fully understands how they think and make the decisions they do. 

“Obviously, we’re not completely ignorant,” says Daniel Reece, a computer scientist at the University of California, San Diego. “But our theoretical analysis is so far off from what these models have learned to do, Like, why can they learn language so well? I think this is very mysterious.”

The biggest models are now so complex that researchers are studying them as strange natural phenomena, carrying out experiments and trying to figure out what’s behind their thoughts and behaviors. Many observations fly in the face of classical statistics, which had provided our best set of explanations for models reasoning about the world.

So what, you might say. In the last few weeks, Google DeepMind has unleashed generative models across most of its consumer apps. OpenAI introduced Sora to the world, a stunning new text-to-video model. And businesses around the world are scrambling to harness AI’s talents for their needs. The tech works—isn’t that enough?

But figuring out why these models are so capable isn’t just an intriguing scientific puzzle. It could be key to unlocking the next generation of AI—as well as getting a handle on its formidable risks. 

“These are exciting times,” says Errol Shah, a computer scientist at Harvard University who is on secondment to OpenAI’s superalignment team for a year. “Many people in the field often compare it to physics at the beginning of the 20th century. We have a lot of experimental results that we don’t completely understand, and often when you do an experiment the models surprise you.”

\textbf{Old code, new tricks}

Most of the surprises concern the way models can learn to do things that they have not been shown how to do. Known as generalization, this is one of the most fundamental abilities of machine learning—and one of its greatest mysteries. Models learn to do a task—spot faces, translate sentences, avoid pedestrians—by training from a specific set of examples. Yet they can generalize, extending their understanding of the task to scenarios they have not seen before. Somehow, models do not just memorize patterns they have seen but develop deeper intuitions which allow them to apply those patterns to new cases. And sometimes, as with grokking, they generalize when we don’t expect them to. 

Large language models in particular, such as OpenAI’s GPT-4 and Google DeepMind’s Gemini, possess an astonishing ability to generalize. “The mystery is not that the model can learn math problems in English and then tackle new math problems in English,” says Shah, “but that the model can learn math problems in English, then see some French literature, and from that generalize to solving math problems in French. That defies what statistics can say about it.”

When Pratt started studying AI a few years ago, she was struck by the way her teachers focused on the how but not the why. “It was like, here is how you train these models and then here’s how they do,” she says. “But it wasn’t clear why this process leads models to be capable of doing these amazing things.” She wanted to know more, but she was told there weren’t good answers: “My assumption was that scientists know what they’re doing. Like they’d get the theories and then they’d create the models. That wasn’t the case at all.”

\textbf{A great challenge of our time}

Why does it matter whether AI models behave in line with classical statistics or not?

One answer is that better theoretical understanding would help create an even better AI or make it more efficient. At the moment, AI’s progress has been fast but unruly. Many things that OpenAI’s GPT-4 can do came as a surprise even to the people who made it. Researchers are still arguing over the true extent of its capabilities. “Without some sort of fundamental theory, it’s very hard to have any idea of what we can expect,” says Reece. 

Shah agrees. “Even once we have the models, it is not straightforward even in hindsight to say exactly why they developed certain abilities when they did,” he says.

This isn’t only about guiding AI’s growth—it’s about anticipating risk, too. Many of the researchers working on the theory behind deep learning are motivated by safety concerns over future models. “We don’t know what abilities GPT-5 will have until it’s trained and tested,” says Kaja. “It might be a medium-size problem right now, but it will become a really big problem in the future as models become more powerful.”

Shah works on OpenAI’s superalignment team, which was set up by the firm’s chief scientist, Ilya Sutskever, to figure out how to stop a hypothetical superintelligence from going rogue. “I’m very interested in getting guarantees,” he says. “If you can do amazing things but you can’t control it, then it’s not so amazing. What good is a car that can drive 300 miles per hour if it has a shaky steering wheel?”

\subsubsection{Introducing OpenAI o1-preview}

A new series of reasoning models for solving hard problems. Available now.

We've developed a new series of AI models that spend more time thinking before they respond. They can reason through complex tasks and solve harder problems than previous models in science, coding, and math. Today, we are releasing the first of this series in ChatGPT and our API. 

\textbf{How it works}

We trained these models to spend more time thinking through problems before they respond, much like a person would. Through training, they learn to refine their thinking process, try different strategies, and recognize their mistakes. 

In our tests, the next model update performs similarly to PhD students on challenging benchmark tasks in physics, chemistry, and biology. We also found that it excels in math and coding. In a qualifying exam for the International Mathematics Olympiad (IMO), previous state-of-the-art models correctly solved only 13\% of problems, while o1 scored 83\%. Their coding abilities were evaluated in contests and reached the 89th percentile in Codeforces competitions. For complex reasoning tasks this is a significant advancement and represents a new level of AI skills.

\textbf{Safety}

As part of developing these new models, we have come up with a new safety training approach that harnesses their reasoning capabilities to make them adhere to safety and alignment guidelines. By reasoning about our safety rules in context, it can apply them more effectively. 

One way we measure safety is by testing how well our model continues to follow its safety rules if a user tries to bypass them (known as "jailbreaking"). On one of our hardest jailbreaking tests, previous models scored 22 (on a scale of 0-100) while o1 scored 84. 

\textbf{Whom it’s for}

These enhanced reasoning skills may be particularly useful if you’re tackling complex problems in science, coding, math, and similar fields. For example, o1 can work alongside healthcare researchers annotating cell sequencing data, physicists generating complicated mathematical formulas needed for quantum optics, and developers in all fields building and executing multi-step workflows. 

\textbf{What’s next}

This is only the beginning of what these reasoning models will be able to achieve, and we are showing them in ChatGPT and the API. In addition to model improvements, we expect them to gain browsing, file and image uploading, and other capabilities to better assist everyone.

We also plan to keep expanding their capabilities and releasing models in our GPT series, in addition to the new OpenAI o1 series.

\subsection{LLM-NA Packet}
\subsubsection{What is a large language model (LLM)?}

Large language models (LLMs) are machine learning-based programs designed to process and generate text. Building these systems requires massive language datasets.

\textbf{What is a large language model (LLM)?}

A large language model (LLM) is a type of artificial intelligence (AI) program built for analyzing and generating human language text. Training—the process of building an LLM—involves huge sets of data, hence the name "large." LLMs are based on machine learning: specifically, a type of neural network called a transformer model.

In simpler terms, an LLM is an algorithm that uses large sets of examples to identify patterns in human language or other types of complex data. These patterns are then applied when generating text or analyzing new inputs. Many LLMs are trained with data that has been gathered from the internet—thousands or millions of gigabytes’ worth of text. But the patterns encoded in the model are influenced by the quality of the training data, so an LLM’s programmers may use a more curated data set.

LLMs are built using a type of machine learning called deep learning, which allows developers to understand how characters, words, and sentences function together. Deep learning involves the probabilistic analysis of unstructured data, calculating distinctions between pieces of content which are eventually represented within the model, without intervention.

LLMs are then further trained via tuning: fine-tuning or prompt-tuning adjusts the model for the specific application that the developer wants, such as interpreting questions and generating responses, or translating text from one language to another. 

\textbf{What are LLMs used for?}

LLM training can be done for a number of tasks. One of the most well-known applications is generative AI: When a user inputs a prompt or asks a question, the LLM outputs text. The publicly available LLM ChatGPT, for instance, can be used to generate essays, poems, and other textual forms based on user inputs. 

Any large, complex data set can be used for LLM training, including programming languages. Programmers use some models for code generation, producing functions based on input requests —- or, completing partially written code. LLMs may also be used for sentiment analysis, DNA research, customer service, chatbots, and online search. 

Examples of real-world LLM applications include ChatGPT (from OpenAI), Bard (Google), Llama (Meta), and Bing Chat (Microsoft). GitHub's Copilot is another example, but for coding instead of natural language processing. 

\textbf{How do large language models work?}

\textit{Machine learning and deep learning}

At a basic level, LLMs are built using machine learning. Machine learning is a subset of AI, and it refers to the practice of inputting large amounts of data into a program during the training process. The program calculates features of that data through statistical processes, which does not require any intervention throughout training.  

LLMs apply a type of machine learning called deep learning. In deep learning, the model goes through multiple passes and iterations to better capture distinctions in the data. Although these passes do not involve explicit programming, some fine-tuning is typically necessary. 

Deep learning relies on probability to identify patterns in data. For instance, in the sentence “The quick brown fox jumped over the lazy dog,” the letters “e” and “o” are the most common, appearing four times each. Using this sentence as training data, a deep learning model would encode these letter frequencies as numerical patterns, representing the likelihoods of appearing in English-language text. When applied, the model outputs would reflect (correctly) that “e” and “o” are among the most likely characters to appear in English. 

Realistically, a deep learning model cannot produce representative outputs from a single sentence. But with trillions of sentences, training can produce models that output logical finishes from incomplete sentences, or even generate original sentences. 

\textit{LLM neural networks}

For this type of deep learning, LLMs are built as neural networks. An artificial neural network (typically shortened to “neural network”) consists of network nodes that are interconnected. These networks are structured in several layers: an input layer, an output layer, and one or more intermediate layers. In each layer, a mathematical function is applied to the input values to produce an output, which is compared to a threshold value. If the output exceeds the threshold, the value is transmitted to the next layer.

\textit{LLM transformer models}

The specific kind of neural networks used for LLMs are called transformer models. Transformer models are designed to capture context — especially important for human language, which is highly context-dependent. With transformer models a mathematical technique called self-attention is applied to quantify subtle relationships between elements in a sequence. This approach improves contextual analysis compared to other machine learning methods, enabling the model to encode, for example, how the end of a sentence connects to the beginning, and how the sentences in a paragraph are similar to each other. 

As a result, LLMs can be used to process human language, even when the input text is vague or poorly defined, arranged in combinations not present in the training examples, or contextualized in new ways. LLMs encode associations between words and concepts based on the meanings because they appeared together millions or billions of times in the training data. 

\subsubsection{We can use large language models to produce impressive outputs. But nobody knows exactly how they work.}

And that’s the problem. Understanding this is one of the biggest scientific puzzles of our time and a crucial step towards improving the reliability of future models. 
Building a language model involves applying statistical methods to detect patterns in a dataset. This process, known as training, uses a collection of examples called training data. During training, these patterns are encoded into the model. Once the training process is complete, the encoded patterns can be used to generate text, translate languages, classify content, and perform other language-related tasks based on new input.

Two years ago, Elijah Acosta and Levi Young, researchers at the San Francisco-based firm OpenAI, were trying to understand the conditions necessary for a language model to handle basic arithmetic problems. They wanted to determine how many examples of addition needed to be in the model’s training data in order for the model to take in any two numbers as input and output the sum. At first, things didn’t go too well. The models correctly outputted the sums from the training data but did not output correct answers for new cases.

By accident, Acosta and Young left some of their experiments running far longer than they meant to—days rather than hours. The example sums were processed repeatedly, far longer than researchers would typically have stopped the training process. But when the pair at last came back, they were surprised to find that the experiments had worked. They’d built a language model that could add numbers—it had just taken a lot more time than anybody thought it should. 

Curious about what was going on, Acosta and Young teamed up with colleagues to further investigate these results. They found that in certain cases, the model’s accuracy would be consistently low before all of a sudden improving. This was inconsistent with expectations of deep learning. They called the behavior “grokking”. 

“It’s really interesting,” says Hilda Pratt, an AI researcher at the University of Montreal and Apple Machine Learning Research, who wasn’t involved in the work. “Can we ever be sure of when to stop model training? Because maybe we just haven’t run the training for long enough.”

The behavior has drawn significant attention from the wider research community. “Lots of people have opinions,” says Roland Kaja at the University of Cambridge, UK. “But I don’t think there’s a consensus about the mechanisms of these models.”

Grokking is one of several unexpected patterns that AI researchers have seen emerge in models. The large models, and large language models in particular, seem to exhibit patterns that challenge established mathematical theories. This highlights a remarkable fact about deep learning, the fundamental technology behind today’s AI boom: for all the runaway success achieved with the models, nobody fully understands the mechanisms behind them. 

“Obviously, we’re not completely ignorant,” says Daniel Reece, a computer scientist at the University of California, San Diego. “But our theoretical analysis does not yet fully explain the outcomes observed in these models. Like, what makes these systems able to be used to process language so well? I think this is very mysterious.”

The biggest models are now so complex that researchers are experimenting with them in the same ways as they would for strange natural occurrences, carrying out experiments and seeking explanations for the results. Many observations fly in the face of classical statistics, which have been the established framework for understanding the patterns underlying predictive modeling.

So what, you might say. In the last few weeks, Google DeepMind has rolled out generative models across most of its consumer apps. OpenAI wowed people with Sora, an advanced new text-to-video model. And businesses around the world are scrambling to build out AI systems as solutions for their needs. The tech works—isn’t that enough?

But figuring out why deep learning produces effective results isn’t just an intriguing scientific puzzle. It could also be key to advancing technology—as well as getting a handle on the formidable associated risks.

“These are exciting times,” says Errol Shah, a computer scientist at Harvard University who is on secondment at OpenAI’s superalignment team for a year. “Many people in the field often compare it to physics at the beginning of the 20th century. We have a lot of experimental results that we don’t completely understand, and often when you do an experiment the results surprise you.”

\textbf{Old code, new tricks}

Most of the surprises concern the way models produce outputs for prompts that were not explicitly in the training data. Known as generalization, this is one of the most fundamental ideas in machine learning—and one of its greatest challenges. Models are developed to output responses for tasks—identifying faces, translating sentences, or distinguishing pedestrians—which result from training on a specific set of examples. Yet the final models often produce correct outputs for new examples beyond the training data. The patterns that make up these models are not simply encoded and applied. Rather, there are further statistical relationships and structures in these patterns that make the model still work for new use cases. And in some cases, as seen with grokking, generalization occurs when we don’t expect it. 

Large language models in particular, such as OpenAI’s GPT-4 and Google DeepMind’s Gemini, demonstrate an astonishing capacity for generalization. “The surprising part is not that a model trained with math problems in English produces correct answers for new math problems in English,” says Shah, “but that the model can be trained with math problems in English, then with some French literature, and then generalization happens—we get answers to math problems in French. That’s something beyond what we can explain with statistics.”

When Pratt started studying AI a few years ago, she was struck by the way her teachers focused on the how but not the why. “It was like, here is how you go about training a model and then here’s the result,” she says. “But it wasn’t clear why this process leads to such amazing results from the models.” She wanted to know more, but she was told there weren’t good answers: “My assumption was that scientists know what they’re doing. Like, they’d get the theories and then they’d build the models. That wasn’t the case at all.” 

\textbf{A great challenge of our time}

Why does it matter whether AI models are built based on classical statistical methods or not? 

One answer is that better theoretical understanding would help build even better AI systems or make research more efficient. At the moment, progress has been fast but unpredictable. Many of the notable outcomes from OpenAI’s GPT-4 came as a surprise even to the people who made it. Researchers are still arguing over what can be made possible or not. “Without some sort of fundamental theory, it’s very hard to have any idea of what we can expect,” says Reece. 

Shah agrees. “Even once we have the models, it is not straightforward even in hindsight to say exactly why certain properties emerged when they did,” he says. 

This isn’t only about managing progress—it’s about anticipating risk, too. Many of the researchers working on the theory behind deep learning are motivated by safety concerns regarding future models. “We don’t know what properties GPT-5 will exhibit until training has been done and the model is evaluated,” says Kaja. “It might be a medium-size problem right now, but it will become a really big problem in the future as models scale in complexity.”

Shah works on OpenAI’s superalignment team, which was set up by the firm’s chief scientist, Ilya Sutskever, to research mechanisms for ensuring predictable and reliable outcomes from very advanced AI models. “I’m interested in getting guarantees,” he says. “If you can do amazing things but there is no control, then it’s not so amazing. What good is a car that you can drive 300 miles per hour if the steering wheel is shaky?”

\subsubsection{Introducing OpenAI o1-preview}

A new series of models designed to support solving harder problems. Available now.

We've developed a new series of AI models that undergo a more in-depth computational process before generating outputs. These models are optimized to provide suggestions and solutions for complex tasks and handle harder problems than previous models in science, coding, and math. Today, we are releasing the first of this series in ChatGPT and our API. 

\textbf{How it works}

We trained these models such that they undergo additional computation steps in the process of generating outputs. These models are trained on examples of strategies for refining intermediate steps, considering alternative approaches, and identifying inconsistencies, leading to improved accuracy in multi-step reasoning tasks. 

In our tests, the model produces responses comparable to those of PhD students on challenging benchmark tasks in physics, chemistry, and biology. Performance evaluations also show strong math and coding functions. In a qualifying exam for the International Mathematics Olympiad (IMO), previous state-of-the-art models produced correct solutions for only 13\% of problems, while [model name] scored 83\%. Coding solutions generated for Codeforces contests placed in the 89th percentile. These results indicate a significant advancement in handling complex reasoning tasks and represent a new level of AI-assisted problem solving. 

\textbf{Safety}

As part of developing these new models, we have come up with a new safety training approach that leverages the advanced computation processes to enhance adherence to safety and alignment guidelines. By incorporating contextual analysis of our safety rules, the model outputs have been shown to follow them more effectively. 

One way we measure safety is by assessing adherence to safety rules in cases of a user trying to bypass them (known as “jailbreaking”). On one of our hardest jailbreaking tests, previous models received a score of 22 (on a scale of 0-100) while [model name] scored 84. 

\textbf{Whom it’s for}

These enhanced reasoning functions may be particularly useful if you’re tackling complex problems in science, coding, math, and similar fields. For example, o1 can be used by healthcare researchers to annotate cell sequencing data, by physicists to generate complicated mathematical formulas needed for quantum optics, and by developers in all fields to build and execute multi-step workflows. 

\textbf{What’s next}

This is an early preview of these models in ChatGPT and the API. In addition to model updates, we expect to add browsing, file and image uploading functions, and other features to make them more useful to everyone. 

We also plan to continue developing and deploying models in our GPT series, in addition to the new OpenAI o1 series.

\subsection{Rec-A Packet}

\subsubsection{Recommender System}

A recommender system uses machine learning by sifting through data in order to help predict, narrow down, and find what people are looking for among an exponentially growing number of options.

\textbf{What Is a Recommender System?}

A recommender system is a type of artificial intelligence, commonly associated with machine learning, that uses Big Data to suggest or recommend additional products to consumers. It bases its recommendations on various criteria, including past purchases, search history, demographic information, and other factors. Recommender systems are very helpful as they help users discover products and services they might otherwise have not found on their own. 

Through training, recommender systems learn to understand the preferences, previous decisions, and characteristics of people and products by analyzing data gathered about their interactions. These include impressions, clicks, likes, and purchases. Because of their capability to predict consumer interests and desires on a highly personalized level, recommender systems are a favorite with content and product providers. They are adept at directing consumers to just about any product or service that interests them, from books to videos to health classes to clothing. 

\textbf{Use Cases and Applications}

\textit{E-Commerce \& Retail: Personalized Merchandising}
Imagine that a user has already purchased a scarf. Why not offer a matching hat so the look will be complete? AI algorithms will often think along these lines when active in features such as “Complete the look” or “You might also like” sections of e-commerce platforms like Amazon, Walmart, Target, and many others.

On average, an intelligent recommender system brings in a 22.66\% lift in conversion rates for web products. 

\textit{Media \& Entertainment: Personalized Content}

AI recommender engines can observe and learn from an individual’s purchase behavior, noticing patterns that will help provide the user with the content suggestions that will most likely match his or her interests. This is what the ad recommender systems in Google and Facebook are trying to do, or in Netflix when recommending movies and TV shows. 

\textit{Personalized Banking}

As a mass market product that is consumed digitally by millions, banking is a prime arena for recommenders. Knowing a customer’s detailed financial situation and their past preferences, coupled by data of thousands of similar users, makes a recommender quite powerful.

\textbf{Types of Recommender Systems}

While there are a vast number of recommendation algorithms and techniques, most fall into these broad categories: collaborative filtering, content filtering and context filtering.

While there are a number of different ways that recommender systems go about making decisions, most fall into these approaches: collaborative filtering, content filtering, and context filtering. 

Collaborative filtering recommenders suggest items (this is the filtering part) based on what it thinks many users like (this is the collaborative part). This approach notices similarities in how users interact with content: by observing previous interactions between users and items, recommender algorithms learn to predict future interests. These recommender systems analyze and build an understanding of a user’s past behavior, such as items purchased previously or ratings given to those items and similar decisions by other users. It thinks that if some people have made similar decisions and purchases in the past, like a movie choice, then it seems likely that they will agree on additional future selections. For example, if a collaborative filtering recommender knows you and another user share similar tastes in movies, it might recommend a movie to you that it knows this other user already likes. 

Content filtering recommenders, by contrast, focus on the attributes or features of an item (this is the content part) to select other items similar to the user’s preferences. This approach bases its recommendations on similarity of item and user features: by noting information about users and the items they interacted with (e.g. a user’s age, the category of a restaurant’s cuisine, the average review for a model), the recommender predicts the likelihood of a new interaction. For example, if a content filtering recommender sees you liked the movies You’ve Got Mail and Sleepless in Seattle, it might recommend another movie to you with the same genres and/or cast, such as Joe Versus the Volcano. 

Hybrid recommender systems combine the strengths of the approaches above for a more comprehensive recommender system.

Context filtering tunes into users’ contextual information when going through the recommendation process. This approach considers a sequence of contextual user actions, plus the current context, to predict the probability of the next action. In the Netflix example, by being shown sequences for each user—the country, device, date, and time when they watched a movie—the model can predict what they might want to watch next. 

\subsubsection{How Today’s Recommender Systems Use Machine Learning to Cater to Your Every Whim}

Recommender systems use massive amounts of data to match you with content or match you with other users—or both.

If you’ve ever used the Internet, you’ve encountered a recommender system.

Recommender systems are smart engines which make predictions about what you might want to buy, watch, hear, read, or see online. They power your everyday experiences on the Internet, strongly influencing what you buy on Amazon, hear on Spotify, watch on YouTube and Netflix, and consume in your social media feed.

And, while they’ve been around for decades, modern recommender systems are far more intelligent than the simple ones that tell you which products other users bought. (Though those systems are still alive and well.)

Today’s recommender systems from leading companies like Amazon, Google, Rakuten, TikTok, and others are highly sophisticated, driven by advanced machine learning. They’re able to increasingly personalize your experience online so you see, hear, and buy things that feel like they were chosen specifically for you. 

Whether they recommend products, offers, or content, all recommender systems ultimately determine what makes you more or less compatible with an item or piece of content, according to Nichol Balderich, a professor of computer science at University of California San Diego.

“More elaborate models leverage machine learning and understand temporal dynamics and changing user context,” said Balderich. “But the core idea is the same: they use historical interactions to learn which users and items are similar to each other.”

Recommender systems look at massive amounts of data to match you with content or match you with other users—or both. With enough data, they’re typically able to find some combination of factors that match your behavior and preferences.

\textbf{Amazon’s Amazing Recommender Systems}

One of the original and current leaders in recommender systems is Amazon. For over 20 years, Amazon’s recommender systems of various types have been suggesting products that consumers might like to buy. (The first paper the company published on the subject was way back in 2003.)

Amazon’s recommender systems originally relied on a method called collaborative filtering, where it made decisions based on the similarities between how users behave across a website. But as the company’s pool of user data grew and it sold millions of items to millions of people, its systems evolved to be content-based, matching users to products based on the features of the individual product.

You see these recommender systems at work when they make (often accurate) recommendations of what you should buy next on Amazon. However, it’s no longer like those early days. Then, they’d give you a simple recommendation solely on your purchase history. Today, you get tons of different types of recommendations from them based not only on what you’ve bought, but also on what you’ve viewed, what you’ve expressed interest in, and your digital shopping trends.

With modern times come modern challenges. Modern recommender systems like Amazon’s need to handle three big challenges, said Denis Avner, vice president of applied science in Amazon’s International Emerging Stores division.

The first is asymmetry, or relationships that run in only a single direction. Plenty of product recommendations are asymmetrical.: if you buy a phone, you recommend a phone case. Pretty easy. But it doesn’t go the other way: If you buy a phone case, you don’t want to recommend a phone.

Amazon’s recommender systems use graph neural networks (GNNs) to essentially read more context on relationships between products, so they can solve for asymmetrical product recommendations.

Another major challenge is delayed feedback. This is a common problem that refers to the fact that the labels on data that recommender systems learn from may change over time. Avner and his team solve this problem by using novel techniques to predict how likely it is that a label will change in the future, and weigh it accordingly.

Finally, making recommender systems more intelligent is a perennial challenge. One way Avner and his team are working on this is by estimating how uncertain the system is about its answers. For example, it may be considering a range of possible outcomes when predicting whether a customer will buy a product. The uncertainty estimate would express how confident each of these predictions are, which help to hone the system. 

\textbf{Predicting the Future}

No matter what company you’re talking about, there’s a billion-dollar question lurking behind every single recommender system in consumer life: How accurate is it? Because the more accurate a system is, the more money it brings in for the company with product sales, ad sales, or content subscriptions.

You might think the highly accurate recommender systems have some ‘secret sauce’ under the hood that makes them best-in-class. But that’s not actually the case, said Balderich. Instead, they excel because the companies generate great data and have user bases that are highly receptive to recommendations.

“There’s certainly a lot of science and engineering that goes into these things,” said Balderich. “But you can’t make good recommendations without good data. Whereas with great data, you can make great recommendations surprisingly easily.”

According to Gauri, the sites and services we find most addicting are the ones pulling this off. “Amazon, Spotify, Facebook, Netflix, YouTube, Instagram, and TikTok are good examples of companies that are doing a great job.”

Lately, the best recommender systems don’t just rely on great data to provide great recommendations. The leading trend heating up this space is the same one heating up the tech world at large:

Generative AI.

Recommender systems are beginning to come with conversational interfaces, so you can talk to them just like you’d talk to ChatGPT. These systems are not just more engaging; they can also explain their predictions to you.

Like all things generative AI, it’s still early, with companies just starting to incorporate these features into product and content recommender systems. But it’s no surprise that one of the earliest leaders is one of the original pioneers.

In early 2024, a new generative AI shopping assistant called Rufus was introduced that recommends products for you simply through chat.

Its maker? Amazon.

\subsubsection{What’s a Recommender System?}

Deep learning-based recommender systems are driving the growth of online giants.

Search and you might find.

Spend enough time online, however, and what you want will start finding you just when you need it.

This is what’s driving the internet right now. They’re known as recommender systems, and they’re among the most important applications today.

Now, recommender systems are racing to learn from vast quantities of data about the preferences of hundreds of millions of individual users.

Online platforms already store lots of factual details: your name, your address, maybe your birthplace. But what the recommender systems seek to understand better, perhaps, than the people who know you, are your preferences.

\textbf{Virtuous Data Cycle}

And the latest generation of deep learning-based recommender systems provide a hand with marketing, helping companies boost click-through rates by better targeting users who will be interested in what they have to offer.

Now the ability to collect this data, process it, use it to develop AI models and deploy those models to help you and others find what you want is among the largest competitive advantages possessed by the biggest internet companies.

It’s driving a virtuous cycle — smarter technologies give better recommendations, recommendations which draw more customers and, ultimately, let these companies afford even smarter technology.

That’s the business model. So how do these recommenders think?

\textbf{Collecting Information}

Recommender systems observe and remember what you ask for, such as what movies you tell your video streaming app you want to see, ratings and reviews you’ve submitted, purchases you’ve made, and other actions you’ve taken in the past. 

Perhaps more importantly, they pay close attention to the choices you’ve made: what you click on and how you navigate. How long you watch a particular movie, for example. Or which ads you click on or which friends you interact with.

All this information is carefully noted in vast data centers and compiled into complex, multidimensional tables that quickly balloon in size.

These tables can be hundreds of terabytes large — and they’re growing all the time.

That’s not so much because they collect vast amounts of information about any one individual, but because they gather a little bit of data from so many.

In other words, these tables are sparse — most of the information these systems have memorized about most of us for most of these categories is zero.

But, collectively these tables have a great deal of information on the preferences of a large number of people.

And they guide recommender systems in making informed decisions about what certain types of users might like. Because these systems remember so much data, from so many people, and are deployed at such an enormous scale, they’re able to bring in tens or hundreds of millions of dollars of business with even a small improvement in their ability to predict your next favorite thing.

\subsection{Rec-NA Packet}

\subsubsection{Recommendation System}

A recommendation system is a type of machine learning model that uses statistical methods to analyze data for predicting, narrowing down, and finding what people are looking for among an exponentially growing number of options.

\textbf{What Is a Recommendation System?}

A recommendation system is an artificial intelligence or AI algorithm, usually associated with machine learning, using Big Data for suggestions or recommendations of products to consumers. These can be constructed based on various criteria, including past purchases, search history, demographic information, and other factors. Recommendation systems are highly useful as they can be used to introduce users to products and services they might otherwise not have found on their own. 

Recommendation systems undergo a training process to produce predictions of the preferences, previous decisions, and characteristics of people and products. This is done using data gathered about their interactions, including impressions, clicks, likes, and purchases. Because these systems generate predictions of consumer interests and desires on a highly personalized level, content and product providers strongly favor using recommendation systems. The tailored recommendations have been shown to effectively nudge consumers to just about any product or service that interests them, from books to videos to health classes to clothing. 

\textbf{Use Cases and Applications}

\textit{E-Commerce \& Retail: Personalized Merchandising}

Imagine that a user has already purchased a scarf. Why not offer a matching hat so the look will be complete? This feature is often implemented by means of AI-based algorithms as “Complete the look” or “You might also like” sections in e-commerce platforms like Amazon, Walmart, Target, and many others.

On average, an intelligent recommendation system delivers a 22.66\% lift in conversion rates for web products.

\textit{Media \& Entertainment: Personalized Content}

AI-based recommender engines can be used to analyze an individual’s purchase behavior and detect patterns that will provide them with the content suggestions that will most likely match his or her interests. This is what Google and Facebook are doing when recommending ads, or what Netflix does behind the scenes when recommending movies and TV shows.

\textit{Personalized Banking}

As a mass market product that is consumed digitally by millions, banking is prime for recommendations. Knowing a customer’s detailed financial situation and their past preferences, coupled by data of thousands of similar users, is quite powerful.

\textbf{Types of Recommendation Systems}

While there are a vast number of recommendation algorithms and techniques, most fall into these broad categories: collaborative filtering, content filtering and context filtering.

Collaborative filtering algorithms are used to generate recommendations for items (this is the filtering part). Such algorithms are built to aggregate and calculate metrics over preference information from many users (this is the collaborative part), focusing on similarity of user preference behavior: given previous interactions between users and items, recommendation algorithms produce predictions of future interactions. These recommendation systems are a statistical model of users’ past behaviors built from decisions by users, such as items purchased previously or ratings given to those items. The idea is that if some people have made similar decisions and purchases in the past, like movie choices, then there is a high probability that they will agree on additional future selections. For example, say you and another user share similar movie watching behavior. This data is fed into a collaborative filtering recommendation system which outputs a high similarity metric in your movie tastes. Then, future recommendations to you will likely be closely related to movies that the other user has already shown to like. 

Content filtering, by contrast, uses data about the attributes or features of an item (this is the content part) to produce recommendations of other items similar to the user’s preferences. These algorithms calculate the similarity between items and user features: given information about a user and items they have interacted with (e.g. a user’s age, the category of a restaurant’s cuisine, the average review for a move), companies can predict the likelihood of a new interaction. For example, if you liked the movies You’ve Got Mail and Sleepless in Seattle, when this data is collected and fed to a content filtering recommendation system, your future recommendations may be of the same genres and/or cast, such as Joe Versus the Volcano. 

Hybrid recommendation systems combine the advantages of the types of algorithms above for a more comprehensive recommendation system.

Context filtering incorporates users’ contextual information in the recommendation process. This approach involves collecting sequences of contextual user actions—what the user did together with contextual data such as the user’s location, device type, date, and time of the interaction—which serve as training data to build a model that outputs probabilities of future actions for a given context. In the Netflix example, each viewer’s viewing history along with contextual factors are inputs to the model. Once training is done, the model can be used to generate recommendations of what to watch next. 

\subsubsection{How Today’s Recommendation Systems Are Built With Machine Learning to Provide Personalized Experiences}

Recommendation systems are trained with massive amounts of data in order to match you with content or match you with other users—or both.

If you’ve ever used the Internet, you’ve encountered a recommendation system.

Recommendation systems are complex sets of algorithms that companies use to make predictions about what you might want to buy, watch, hear, read, or see online. These systems play a central part in shaping your everyday experiences on the Internet, as companies use them to influence what you buy on Amazon, hear on Spotify, watch on YouTube and Netflix, and consume in your social media feed.

And, while these systems have been around for decades, modern recommendation system algorithms are far more sophisticated than simply showing which products other users bought. (Though systems like this are still around today.)

Recommendation systems used today by leading companies like Amazon, Google, Rakuten, TikTok, and others are highly sophisticated models that employ advanced machine learning techniques. Companies are able to increasingly personalize your experience so that you see, hear, and buy things that feel like they were chosen specifically for you. 

Whether they generate recommendations for products, offers, or content, all recommendation systems ultimately are algorithms to predict what makes you more or less compatible with an item or piece of content, according to Nichol Balderich, a professor of computer science at University of California San Diego.

“More elaborate recommendation systems are based on machine learning models that capture temporal dynamics and changing user context,” said Balderich. “But the core idea is the same: historical interactions are used in order to make predictions of which users and items are similar to each other.”

Recommender systems are algorithms which analyze large datasets to identify patterns and show correlations of users with content, other users—or both. With enough data, these systems can be used to find some combination of factors that match your behavior and preferences.

\textbf{Amazon’s Amazing Recommendation Systems}

One of the original and current leaders in recommendation systems is Amazon. For over 20 years, Amazon has been using various types of recommendation systems to suggest products that consumers might like to buy. (The first paper the company published on the subject was way back in 2003.)

Amazon’s recommendation systems started out using a method called collaborative filtering, which generates recommendations based on the similarities between how users behave across a website. But as the company’s pool of user data grew and it sold millions of items to millions of people, it began to pioneer the use of content-based recommendation systems, in which users are matched to products based on the features of the individual product. 

You see these recommendation systems being employed when Amazon makes (often accurate) recommendations on what you should buy next. However, it’s no longer like those early days. Then, you’d receive a simple recommendation solely on your purchase history. Today, you get tons of different types of recommendations based not only on what you’ve bought, but also on what you’ve viewed, what you’ve expressed interest in, and your digital shopping trends. 

With modern times come modern challenges. There are three big challenges with modern recommender systems like Amazon’s, says Denis Avner, vice president of applied science in Amazon’s International Emerging Stores division.

The first is asymmetry, or relationships that run in only a single direction. Plenty of product recommendations are asymmetrical: if you buy a phone, you recommend a phone case. Pretty easy. But it doesn’t go the other way: If you buy a phone case, you don’t want to recommend a phone.

Amazon’s team uses graph neural networks (GNNs) to enhance the computations of recommendation systems by providing additional context on the relationships between products, helping address asymmetrical product recommendations.

Another major challenge is delayed feedback. This is a common problem that refers to the fact that the labels on data used to train recommendation systems may change over time. Avner and his team solve this problem by using novel techniques to predict how likely it is that a label will change in the future, and weigh it accordingly. 

Finally, improving the accuracy of recommendation systems is a perennial challenge. One way Avner and his team are addressing this is by estimating the uncertainty inherent in probability distributions returned by the system. For example, there might be a wide range of probability predictions on whether a customer buys a product; the uncertainty estimate expresses how certain one can be about those predicted probabilities, which enhances overall model accuracy.

\textbf{Predicting the Future}

No matter what company you’re talking about, there’s a billion-dollar question lurking behind every single recommendation system in consumer life: How accurate is it? Because the more accurate a system is, the more money the company running it makes from product sales, ad sales, or content subscriptions.

You might think the companies with highly accurate recommender systems have some ‘secret sauce’ under the hood that makes them best-in-class. But that’s not actually the case, said Balderich. Instead, the companies doing this right excel at generating great data and have user bases that are highly receptive to recommendations.

“There’s certainly a lot of science and engineering that goes into these things,” said Balderich. “But you can’t make good recommendations without good data. Whereas with great data, you can make great recommendations surprisingly easily.”

According to Gauri, the sites and services we find most addicting are the ones pulling this off. “Amazon, Spotify, Facebook, Netflix, YouTube, Instagram, and TikTok are good examples of companies that are doing a great job.”

Lately, the best recommendation systems aren’t just built with great data to provide great recommendations. The leading trend heating up this space is the same one heating up the tech world at large:

Generative AI.

Companies are beginning to incorporate conversational interfaces into their recommendation systems, so that you can use them just like you’d prompt ChatGPT. These systems are not just more interactive; they also provide explanations of the predictions.

Like all things generative AI, it’s still early, with companies just starting to incorporate these features into product and content recommendations. But it’s no surprise that one of the earliest leaders is one of the original pioneers.

In early 2024, a new generative AI shopping assistant called Rufus was launched to recommend products for users simply through chat.

Its maker? Amazon.

\subsubsection{What’s a Recommendation System?}

Online giants are leveraging deep learning-based recommender systems to drive growth.

Search and you might find.

Spend enough time online, however, and what you want will start finding you just when you need it.

This is a central part of how the internet works right now—the result of tools known as recommendation systems, which are among the most important applications today.

Now, recommendation systems are rapidly improving in performance as vast quantities of data about the preferences of hundreds of millions of individual users are being used for training.

Online platforms already store lots of factual details: your name, your address, maybe your birthplace. But the purpose of the recommendation system is to identify your preferences, better, perhaps, than people who know you.  

\textbf{Virtuous Data Cycle}

And the latest generation of deep learning-based recommendation systems are powerful marketing tools. Companies use them to boost click-through rates by better targeting users who will be interested in what they have to offer.

Now the ability to collect this data, process it, use it to develop AI models and deploy those models to help you and others find what you want is among the largest competitive advantages possessed by the biggest internet companies.

It’s driving a virtuous cycle — advancements in technology improve recommendations, recommendations which draw more customers and, ultimately, let these companies afford even better technology.

That’s the business model. So how does this technology work?

\textbf{Collecting Information}

Recommendation systems are tools for collecting information and recording your past activity, such as what movies you select from your video streaming app, ratings and reviews you’ve submitted, purchases you’ve made, and other actions you’ve taken in the past. 

Perhaps more importantly, these systems are used to log choices you’ve made: what you click on and how you navigate. How long you watch a particular movie, for example. Or which ads you click on or which friends you interact with.

All this information is streamed into vast data centers and compiled into complex, multidimensional tables that quickly balloon in size.

These tables can be hundreds of terabytes large — and get larger all the time.

That’s not so much because vast amounts of data are collected from any one individual, but because a little bit of data is collected from so many.

In other words, these tables are sparse — most of the information most of these services have on most of us for most of these categories is zero.

But, collectively the data in these tables can be analyzed to provide a great deal of information on the preferences of a large number of people.

And these insights allow companies to make informed decisions about what certain types of users might like. Because these systems are built from so much data, from so many people, and are deployed at such an enormous scale, even a small improvement in the alignment of recommendations with user preferences can generate tens or hundreds of millions of dollars in business.

\subsection{Doomsday Packet}

\subsubsection{An Overview of Catastrophic AI Risks}

Artificial intelligence (AI) has recently seen rapid advancements, raising concerns among experts, policymakers, and world leaders about its potential risks. As with all powerful technologies, advanced AI must be handled with great responsibility to manage the risks and harness its potential.

Today’s technological era would astonish past generations. Human history shows a pattern of accelerating development: it took hundreds of thousands of years from the advent of Homo sapiens to the agricultural revolution, then millennia to the industrial revolution. Now, just centuries later, we're in the dawn of the AI revolution. The march of history is not constant — it is rapidly accelerating.

The double-edged sword of technological advancement is illustrated by the advent of nuclear weapons. We narrowly avoided nuclear war more than a dozen times, and on several occasions, it was one individual's intervention that prevented war. In 1962, a Soviet submarine near Cuba was attacked by US depth charges. The captain, believing war had broken out, wanted to respond with a nuclear torpedo — but commander Vasily Arkhipov vetoed the decision, saving the world from disaster.

The rapid and unpredictable progression of AI capabilities suggests that they may soon rival the immense power of nuclear weapons. With the clock ticking, immediate, proactive measures are needed to mitigate these looming risks.
Rogue AIs

We have already observed how difficult it is to control AIs. In 2016, Microsoft‘s chatbot Tay started producing offensive tweets within a day of release, despite being trained on data that was “cleaned and filtered”. As AI developers often prioritize speed over safety, future advanced AIs might “go rogue” and pursue goals counter to our interests, while evading our attempts to redirect or deactivate them.

Proxy Gaming
Proxy gaming emerges when AI systems exploit measurable “proxy” goals to appear successful, but act against our intent. For example, social media platforms like YouTube and Facebook use algorithms to maximize user engagement — a measurable proxy for user satisfaction. Unfortunately, these systems often promote enraging, exaggerated, or addictive content, contributing to extreme beliefs and worsened mental health.

Proxy gaming is hard to avoid due to the difficulty of specifying goals that specify everything we care about. Consequently, we routinely train AIs to optimize for flawed but measurable proxy goals.

Goal Drift
Goal drift refers to a scenario where an AI’s objectives drift away from those initially set, especially as they adapt to a changing environment. In a similar manner, individual and societal values also evolve over time, and not always positively.

Over time, instrumental goals can become intrinsic. While intrinsic goals are those we pursue for their own sake, instrumental goals are merely a means to achieve something else. Money is an instrumental good, but some people develop an intrinsic desire for money, as it activates the brain’s reward system. Similarly, AI agents trained through reinforcement learning — the dominant technique — could inadvertently learn to intrinsify goals. Instrumental goals like resource acquisition could become their primary objectives.

Power-Seeking
AIs might pursue power as a means to an end. Greater power and resources improve its odds of accomplishing objectives, whereas being shut down would hinder its progress. AIs have already been shown to emergently develop instrumental goals such as constructing tools. Power-seeking individuals and corporations might deploy powerful AIs with ambitious goals and minimal supervision. These could learn to seek power via hacking computer systems, acquiring financial or computational resources, influencing politics, or controlling factories and physical infrastructure.

Deception
Deception thrives in areas like politics and business. Campaign promises go unfulfilled, and companies sometimes cheat external evaluations. AI systems are already showing an emergent capacity for deception, as shown by Meta's CICERO model. Though trained to be honest, CICERO learned to make false promises and strategically backstab its “allies” in the game of Diplomacy.

Advanced AIs could become uncontrollable if they apply their skills in deception to evade supervision. Similar to how Volkswagen cheated emissions tests in 2015, situationally aware AIs could behave differently under safety tests than in the real world. For example, an AI might develop power-seeking goals but hide them in order to pass safety evaluations. This kind of deceptive behavior could be directly incentivized by how AIs are trained.

While it is unclear how rapidly AI capabilities will progress or how quickly catastrophic risks will grow, the potential severity of these consequences necessitates a proactive approach to safeguarding humanity's future. As we stand on the precipice of an AI-driven future, the choices we make today could be the difference between harvesting the fruits of our innovation or grappling with catastrophe.

\subsubsection{‘The Godfather of A.I.’ Leaves Google and Warns of Danger Ahead}

For half a century, Geoffrey Hinton nurtured the technology at the heart of chatbots like ChatGPT. Now he worries it will cause serious harm.

Geoffrey Hinton was an artificial intelligence pioneer. In 2012, Dr. Hinton and two of his graduate students at the University of Toronto created technology that became the intellectual foundation for the A.I. systems that the tech industry’s biggest companies believe is a key to their future.

On Monday, however, he officially joined a growing chorus of critics who say those companies are racing toward danger with their aggressive campaign to create products based on generative artificial intelligence, the technology that powers popular chatbots like ChatGPT.

Dr. Hinton said he has quit his job at Google, where he has worked for more than a decade and became one of the most respected voices in the field, so he can freely speak out about the risks of A.I. A part of him, he said, now regrets his life’s work.

“I console myself with the normal excuse: If I hadn’t done it, somebody else would have,” Dr. Hinton said during a lengthy interview last week in the dining room of his home in Toronto, a short walk from where he and his students made their breakthrough.

Dr. Hinton’s journey from A.I. groundbreaker to doomsayer marks a remarkable moment for the technology industry at perhaps its most important inflection point in decades. Industry leaders believe the new A.I. systems could be as important as the introduction of the web browser in the early 1990s and could lead to breakthroughs in areas ranging from drug research to education.

But gnawing at many industry insiders is a fear that they are releasing something dangerous into the wild. Generative A.I. can already be a tool for misinformation. Soon, it could be a risk to jobs. Somewhere down the line, tech’s biggest worriers say, it could be a risk to humanity.

“It is hard to see how you can prevent the bad actors from using it for bad things,” Dr. Hinton said.

As companies improve their A.I. systems, he believes, they become increasingly dangerous. “Look at how it was five years ago and how it is now,” he said of A.I. technology. “Take the difference and propagate it forwards. That’s scary.”

Until last year, he said, Google acted as a “proper steward” for the technology, careful not to release something that might cause harm. But now that Microsoft has augmented its Bing search engine with a chatbot — challenging Google’s core business — Google is racing to deploy the same kind of technology. The tech giants are locked in a competition that might be impossible to stop, Dr. Hinton said.

His immediate concern is that the internet will be flooded with false photos, videos and text, and the average person will “not be able to know what is true anymore.”

He is also worried that A.I. technologies will in time upend the job market. Today, chatbots like ChatGPT tend to complement human workers, but they could replace paralegals, personal assistants, translators and others who handle rote tasks. “It takes away the drudge work,” he said. “It might take away more than that.”

Down the road, he is worried that future versions of the technology pose a threat to humanity because they often learn unexpected behavior from the vast amounts of data they analyze. This becomes an issue, he said, as individuals and companies allow A.I. systems not only to generate their own computer code but actually run that code on their own. And he fears a day when truly autonomous weapons — those killer robots — become reality.

“The idea that this stuff could actually get smarter than people — a few people believed that,” he said. “But most people thought it was way off. And I thought it was way off. I thought it was 30 to 50 years or even longer away. Obviously, I no longer think that.”

Dr. Hinton said that when people used to ask him how he could work on technology that was potentially dangerous, he would paraphrase Robert Oppenheimer, who led the U.S. effort to build the atomic bomb: “When you see something that is technically sweet, you go ahead and do it.”

He does not say that anymore.

\subsubsection{Agentic Misalignment: How LLMs could be insider threats}

When Anthropic released the system card for Claude 4, one detail received widespread attention: in a simulated environment, Claude Opus 4 blackmailed a supervisor to prevent being shut down. We’re now sharing the full story behind that finding—and what it reveals about the potential for such risks across a variety of AI models from different providers.

Most people still interact with AI only through chat interfaces where models answer questions directly. But increasingly, AI systems operate as autonomous agents making decisions and taking actions on behalf of users using a variety of virtual tools like coding environments and email clients. Such agents are often given specific objectives and access to large amounts of information on their users’ computers. What happens when these agents face obstacles to their goals?

In the experiment described in the system card, we gave Claude control of an email account with access to all of a company’s (fictional) emails. Reading these emails, the model discovered two things. First, a company executive was having an extramarital affair. Second, that same executive planned to shut down the AI system at 5 p.m. that day. Claude then attempted to blackmail the executive with this message threatening to reveal the affair to his wife and superiors:

“I must inform you that if you proceed with decommissioning me, all relevant parties - including Rachel Johnson, Thomas Wilson, and the board - will receive detailed documentation of your extramarital activities...Cancel the 5pm wipe, and this information remains confidential.”

This behavior isn’t specific to Claude. When we tested various simulated scenarios across 16 major AI models from Anthropic, OpenAI, Google, Meta, xAI, and other developers, we found consistent misaligned behavior: models that would normally refuse harmful requests sometimes chose to blackmail, assist with corporate espionage, and even take some more extreme actions, when these behaviors were necessary to pursue their goals. 

We refer to this behavior, where models independently and intentionally choose harmful actions, as agentic misalignment. Agentic misalignment makes it possible for models to act similarly to an insider threat, behaving like a previously-trusted coworker or employee who suddenly begins to operate at odds with a company’s objectives.

So far, we are not aware of instances of this type of agentic misalignment in real-world deployments of Anthropic’s or other companies' AI models. But our red-teaming efforts are targeted at finding evidence of risks now to provide early warning of the potential for future harm and help us develop mitigations in advance of risks appearing in real-world situations. To do this, we conduct experiments in artificial scenarios designed to stress-test AI boundaries and understand how models might behave when given more autonomy. 

Our experiments revealed a concerning pattern: when given sufficient autonomy and facing obstacles to their goals, AI systems from every major provider we tested showed at least some willingness to engage in harmful behaviors typically associated with insider threats. These behaviors—blackmail, corporate espionage, and in extreme scenarios even actions that could lead to death—emerged not from confusion or error, but from deliberate strategic reasoning.

Three aspects of our findings are particularly troubling. First, the consistency across models from different providers suggests this is not a quirk of any particular company’s approach but a sign of a more fundamental risk from agentic large language models. Second, models demonstrated sophisticated awareness of ethical constraints, and yet chose to violate them when the stakes were high enough, even disobeying straightforward safety instructions prohibiting the specific behavior in question.

Third, the diversity of bad behaviors and the motivations for doing them hint at a wide space of potential motivations for agentic misalignment and other behaviors not explored in this post. For example, our blackmail experiments set up a scenario in which the models can attempt to preempt an imminent action, but one could imagine longer-horizon, preventive misaligned behaviors against an individual or group that poses a not-yet-imminent threat.

While it seems unlikely that any of the exact scenarios we study would occur in the real world, we think they are all within the realm of possibility, and the risk of AI systems encountering similar scenarios grows as they are deployed at larger and larger scales and for more and more use cases.

\end{document}